\documentclass{article}

\usepackage[frozencache]{minted}

\usepackage{times}
\usepackage{graphicx} 
\usepackage{caption}
\usepackage{subcaption}

\usepackage{natbib}

\usepackage{algorithm}
\usepackage{algorithmic}

\usepackage[utf8]{inputenc}
\usepackage{amsmath, bm}
\usepackage{amsfonts}
\usepackage{amssymb}
\usepackage{graphicx}

\DeclareMathOperator*{\argmin}{arg\,min}

\usepackage{algorithm}
\usepackage{algorithmic}

\usepackage{listings}

\newcommand{\ica}{\hspace{0.25cm}}

\usepackage{enumitem}

\usepackage{capt-of}

\newcommand{\bo}{\omega}
\newcommand{\KL}{\text{KL}}

\newcommand{\softmax}{\text{Softmax}}

\newcommand{\logsumexp}{\text{log-sum-exp}}

\newcommand{\N}{\mathcal{N}}

\newcommand{\td}{\text{d}}
\newcommand{\f}{\mathbf{f}}
\newcommand{\x}{\mathbf{x}}
\newcommand{\Bb}{\mathbf{b}}

\newcommand{\y}{\mathbf{y}}

\newcommand{\W}{\mathbf{W}}

\newcommand{\X}{\mathbf{X}}
\newcommand{\Y}{\mathbf{Y}}

\newcommand{\I}{\mathbf{I}}
\newcommand{\M}{\mathbf{M}}

\newcommand{\bz}{\mathbf{0}}
\newcommand{\bepsilon}{\text{\boldmath$\epsilon$}}

\newcommand{\boh}{\widehat{\text{\boldmath$\omega$}}}

\allowdisplaybreaks

\usepackage{hyperref}

\usepackage[accepted]{icml2017}

\icmltitlerunning{Dropout Inference in Bayesian Neural Networks with Alpha-divergences}

\begin{document} 

\twocolumn[
\icmltitle{Dropout Inference in Bayesian Neural Networks with Alpha-divergences}




\begin{icmlauthorlist}
\icmlauthor{Yingzhen Li}{cam}
\icmlauthor{Yarin Gal}{cam,ati}
\end{icmlauthorlist}

\icmlaffiliation{cam}{University of Cambridge, UK}
\icmlaffiliation{ati}{The Alan Turing Institute, UK}

\icmlcorrespondingauthor{Yingzhen Li}{yl494@cam.ac.uk}

\icmlkeywords{boring formatting information, machine learning, ICML}

\vskip 0.3in
]



\printAffiliationsAndNotice{}  

\begin{abstract} 

To obtain uncertainty estimates with real-world Bayesian deep learning models, practical inference approximations are needed. Dropout variational inference (VI) for example has been used for machine vision and medical applications, but VI can severely underestimates model uncertainty. Alpha-divergences are alternative divergences to VI's KL objective, which are able to avoid VI's uncertainty underestimation. But these are hard to use in practice: existing techniques can only use Gaussian approximating distributions, and require existing models to be changed radically, thus are of limited use for practitioners. We propose a re-parametrisation of the alpha-divergence objectives, deriving a simple inference technique which, together with dropout, can be easily implemented with existing models by simply changing the loss of the model. We demonstrate improved uncertainty estimates and accuracy compared to VI in dropout networks. We study our model's epistemic uncertainty far away from the data using adversarial images, showing that these can be distinguished from non-adversarial images by examining our model's uncertainty.
\end{abstract} 

\vspace{-8mm}
\section{Introduction}
Deep learning models have been used to obtain state-of-the-art results on many tasks \citep{krizhevsky2012imagenet, szegedy2014going, sutskever2014sequence, sundermeyer2012lstm, mikolov2010recurrent, kalchbrenner2013recurrent}, and in many pipelines these models have replaced the more traditional \textit{Bayesian probabilistic} models \citep{sennrich2016Edinburgh}. 
But unlike deep learning models, Bayesian probabilistic models can capture parameter uncertainty and its induced effects over predictions, capturing the models' ignorance about the world, and able to convey their increased uncertainty on out-of-data examples. This information can be used, for example, to identify when a vision model is given an adversarial image (studied below), or to tackle many problems in AI safety \citep{amodei2016concrete}. With model uncertainty at hand, applications as far-reaching as safety in self-driving cars can be explored, using models which can propagate their uncertainty up the decision making pipeline \citep{Gal2016Uncertainty}. With deterministic deep learning models this invaluable uncertainty information is often lost.

Bayesian deep learning -- an approach to combining Bayesian probability theory together with deep learning -- allows us to use state-of-the-art models and at the same time obtain model uncertainty \citep{Gal2016Uncertainty, Gal2016Bayesian}. Originating in the 90s \citep{neal1995bayesian, mackay1992practical, denker1991transforming}, Bayesian neural networks (BNNs) in particular have started gaining in popularity again \citep{graves2011practical, blundell2015weight, hernandez2015probabilistic}. BNNs are standard neural networks (NNs) with prior probability distributions placed over their weights. Given observed data, inference is then performed to find what are the more likely and less likely weights to explain the data. But as easy it is to formulate BNNs, is as difficult to perform inference in them. Many approximations have been proposed over the years \citep{denker1991transforming, neal1995bayesian, graves2011practical, blundell2015weight, hernandez2015probabilistic, hernandez2016black}, some more practical and some less practical.
A practical approximation for inference in Bayesian neural networks should be able to scale well to large data and complex models (such as convolutional neural networks (CNNs) \citep{rumelhart1985learning, lecun1989backpropagation}). Much more important perhaps, it would be impractical to change existing model architectures that have been well studied, and it is often impractical to work with complex and cumbersome techniques which are difficult to explain to non-experts. Many existing approaches to obtain model confidence often do not scale to complex models or large amounts of data, and require us to develop new models for existing tasks for which we already have well performing tools \citep{Gal2016Uncertainty}. 

One possible solution for practical inference in BNNs is variational inference (VI) \citep{jordan1999introduction}, a ubiquitous technique for approximate inference. 
Dropout variational distributions in particular (a mixture of two Gaussians with small standard deviations, and with one component fixed at zero) can be used to obtain a practical inference technique \citep{gal2016dropout}. 
These have been used for machine vision and medical applications \citep{kendall2016modelling, kendall2015bayesian, angermueller2015multi, yang2016fast}.
Dropout variational inference can be implemented by adding dropout layers \citep{hinton2012improving, srivastava2014dropout} before every weight layer in the NN model. Inference is then carried out by Monte Carlo (MC) integration over the variational distribution, in practice implemented by simulating stochastic forward passes through the model at test time (referred to as MC dropout). Although dropout VI is a practical technique for approximate inference, it also has some major limitations. Dropout VI can severely underestimate model uncertainty \citep[Section 3.3.2]{Gal2016Uncertainty} -- a property many VI methods share \citep{turner2011two}. This can lead to devastating results in applications that \textit{must rely} on good uncertainty estimates such as AI safety applications. 

Alternative objectives to VI’s objective are therefore needed. Black-box $\alpha$-divergence minimisation \citep{hernandez2016black, li2016variational, minka2005divergence} is a class of approximate inference methods extending on VI, approximating EP’s energy function \cite{minka2001ep} as well as the Hellinger distance \cite{hellinger1909neue}. These were proposed as a solution to some of the difficulties encountered with VI. However, the main difficulty with $\alpha$-divergences is that the divergences are hard to use in practice. Existing inference techniques only use Gaussian approximating distributions, with the density over the approximation having to be evaluated explicitly many times.
The objective offers a limited intuitive interpretation which is difficult to explain to non-experts, and of limited use for engineers \citep[Section 2.2.2]{Gal2016Uncertainty}. Perhaps more important, current $\alpha$-divergence inference techniques require existing models and code-bases to be changed radically to perform inference in the Bayesian counterpart to these models. To implement a complex CNN structure with the inference and code of  \citep{hernandez2016black}, for example, one would be required to re-implement many already-implemented software tools.

In this paper we propose a re-parametrisation of the induced $\alpha$-divergence objectives, and by relying on some mild assumptions (which we justify below), derive a simple approximate inference technique which can easily be implemented with existing models. Further, we rely on the dropout approximate variational distribution and demonstrate how inference can be done in a practical way -- requiring us to \textit{only change the loss of the NN, $\mathcal{L}(\theta)$, and to perform multiple stochastic forward passes at training time}. In particular, given $l(\cdot, \cdot)$ some standard NN loss such as cross entropy or the Euclidean loss, and $\{ \f^{\boh_k}(\bm{x}_n) \}_{k=1}^K$ a set of $K$ stochastic dropout network outputs on input $\x_n$ with randomly masked weights $\boh_k$, our proposed objective is:
\begin{align*}
\mathcal{L}(\theta) &= - \frac{1}{\alpha} \sum_n \logsumexp \left[-\alpha \cdot l(y_n, \f^{\boh_k}(\bm{x}_n)) \right]
+ L_2(\theta)
\end{align*}
with $\alpha$ a real number, $\theta$ the set of network weights to be optimised, and an $L_2$ regulariser over $\theta$.
By selecting $\alpha=1$ this objective directly optimises the per-point predictive log-likelihood, while picking $\alpha \rightarrow 0$ would focus on increasing the training accuracy, recovering VI. 

Specific choices of $\alpha$ will result in improved uncertainty estimates (and accuracy) compared to VI in dropout BNNs, without slowing convergence time. We demonstrate this through a myriad of applications, including an assessment of fully connected NNs in regression and classification, and an assessment of Bayesian CNNs. Finally, we study the uncertainty estimates resulting from our approximate inference technique. We show that our models' uncertainty increases on adversarial images generated from the MNIST dataset, suggesting that these lie outside of the training data distribution. This in practice allows us to tell-apart such adversarial images from non-adversarial images by examining epistemic model uncertainty. 

\section{Background}
We review background in Bayesian neural networks and approximate variational inference. In the next section we discuss $\alpha$-divergences.

\subsection{Bayesian Neural Networks} \label{sec:BNN}

Given training inputs $\X = \{ \x_1, \hdots, \x_N \}$ and their corresponding outputs $\Y = \{\y_1, \hdots, \y_N\}$, in parametric Bayesian regression we would like to infer a distribution over parameters $\bo$ of a function $\y = \f^\bo(\mathbf{x})$ that could have generated the outputs. 
Following the Bayesian approach, to find parameters that could have generated our data, we put some \textit{prior} distribution over the space of parameters $p_0(\bo)$. This distribution captures our prior belief as to which parameters are likely to have generated our outputs before observing any data.
We further need to define a probability distribution over the outputs given the inputs $p(\y | \x, \bo)$.
For classification tasks we assume a softmax likelihood, 
\begin{align*}
p \big( y | \x, \bo \big) = \softmax \left( \f^\bo(\x) \right)
\end{align*}
or a Gaussian likelihood for regression. 
Given a dataset $\X, \Y$, we then look for the \textit{posterior} distribution over the space of parameters: $p(\bo | \X, \Y)$.
This distribution captures how likely the function parameters are, given our observed data.
With it we can predict an output for a new input point $\x^*$ by integrating
\begin{align} \label{eq:post}
p(\y^* | \x^*, \X, \Y) = \int p(\y^* | \x^*, \bo) p(\bo | \X, \Y) \td \bo.
\end{align}

One way to define a distribution over a parametric set of functions is to place a prior distribution over a \textit{neural network's} weights $\bo = \{ \W_i \}_{i=1}^L$, resulting in a \textit{Bayesian NN} \citep{mackay1992practical,neal1995bayesian}.
Given weight matrices $\W_i$ and bias vectors $\Bb_i$ for layer $i$, we often place standard matrix Gaussian prior distributions over the weight matrices, $p_0(\W_i) = \N(\W_i; \bz, \I)$
and often assume a point estimate for the bias vectors for simplicity.

\subsection{Approximate Variational Inference in Bayesian Neural Networks}
In approximate inference, we are interested in finding the distribution of weight matrices (parametrising our functions) that have generated our data. This is the posterior over the weights given our observables $\X, \Y$: $p( \bo | \X, \Y )$, which is not tractable in general. Existing approaches to approximate this posterior are through \textit{variational inference} (as was done in \citet{hinton1993keeping,
barber1998ensemble,
graves2011practical,
blundell2015weight}). We need to define an approximating variational distribution $q_\theta( \bo )$ (parametrised by variational parameters $\theta$), and then minimise w.r.t. $\theta$ the KL divergence \citep{kullback1951information, kullback1959information} between the approximating distribution and the full posterior:
\begin{align} \label{eq:KL:BNN}
\KL \big( q_\theta(\bo) || p(\bo | \X, \Y ) \big)
&\propto - \int q_\theta(\bo) \log p(\Y | \X, \bo) \td \bo \notag \\
&\qquad + \KL(q_\theta(\bo) || p_0(\bo)) 
\notag \\
&=- \sum_{i=1}^N \int q_\theta(\bo) \log p(\y_i | \f^\bo(\x_i)) \td \bo \notag \\
&\qquad + \KL(q_\theta(\bo) || p_0(\bo)),
\end{align}
where $A \propto B$ is slightly abused here to denote equality up to an additive constant (w.r.t.\ variational parameters $\theta$).

\subsection{Dropout Approximate Inference}

Given a (deterministic) neural network, stochastic regularisation techniques in the model (such as dropout \citep{hinton2012improving, srivastava2014dropout}) can be interpreted as variational Bayesian approximations in a Bayesian NN with the same network structure \citep{gal2016dropout}. This is because applying a stochastic regularisation technique is equivalent to multiplying the NN weight matrices $\M_i$ by some random noise $\bepsilon_i$ (with a new noise realisation for each data point). The resulting stochastic weight matrices $\W_i = \bepsilon_i \M_i$ can be seen as draws from the approximate posterior over the BNN weights, replacing the deterministic NN's weight matrices $\M_i$.
Our set of variational parameters is then the set of matrices $\theta = \{ \M_i \}_{i=1}^L$.
For example, dropout can be seen as an approximation to Bayesian NN inference with \textit{dropout approximating distributions}, where the rows of the matrices $\W_i$ distribute according to a mixture of two Gaussians with small variances and the mean of one of the Gaussians fixed at zero. 
The uncertainty in the weights induces prediction uncertainty by marginalising over the approximate posterior using Monte Carlo integration:
\begin{align*}
p(y=c | \x, \X, \Y) &= \int p(y=c | \x, \bo) p(\bo | \X, \Y) \td \bo \\
&\approx 
\int p(y=c | \x, \bo) q_\theta(\bo) \td \bo \\
&\approx 
\frac{1}{K} \sum_{k=1}^K p(y=c | \x, \boh_k)
\end{align*}
with $\boh_k \sim q_\theta(\bo)$, where $q_\theta(\bo)$ is the Dropout distribution \citep{Gal2016Uncertainty}.
Given its popularity, we concentrate on the dropout stochastic regularisation technique throughout the rest of the paper, although any other stochastic regularisation technique could be used instead (such as multiplicative Gaussian noise \citep{srivastava2014dropout} or dropConnect \citep{wan2013regularization}).

Dropout VI is an example of practical approximate inference, but it also underestimates model uncertainty \citep[Section 3.3.2]{Gal2016Uncertainty}. This is because minimising the KL divergence between $q(\bo)$ and $p(\bo|\X, \Y)$ penalises $q(\bo)$ for placing probability mass where $p(\bo|\X, \Y)$ has no mass, but does not penalise $q(\bo)$ for not placing probability mass at locations where $p(\bo|\X, \Y)$ \textit{does have mass}.
We next discuss $\alpha$-divergences as an alternative to the VI objective.

\section{Black-box $\alpha$-divergence minimisation}
In this section we provide a brief review of the \emph{black box alpha} (BB-$\alpha$, \citet{hernandez2016black}) method upon which the main derivation in this paper is based. Consider approximating the following distribution:
$$p(\bo) = \frac{1}{Z} p_0(\bo) \prod_{n} f_n(\bo).$$
In Bayesian neural networks context, these factors $f_n(\bo)$ represent the likelihood terms $p(\y_n | \x_n, \bo)$, $Z = p(\Y | \X)$, and the approximation target $p(\bo)$ is the exact posterior $p(\bo | \X, \Y)$. 
Popular methods of approximate inference include variational inference (VI) \cite{jordan1999introduction} and expectation propagation (EP) \cite{minka2001ep}, where these two algorithms are special cases of power EP \cite{minka2004powerep} that minimises Amari's $\alpha$-divergence \cite{amari1985book} $\mathrm{D}_{\alpha}[p||q]$ in a \emph{local} way:
\begin{equation*}
\label{eq:alpha_divergence}
\mathrm{D}_{\alpha}[p||q] = \frac{1}{\alpha (1 - \alpha)} \left(1 - \int p(\bo)^{\alpha} q(\bo)^{1 - \alpha} d\bo \right).
\end{equation*}

We provide details of $\alpha$-divergences and local approximation methods in the appendix, and in the rest of the paper we consider three special cases in this rich family:

1. Exclusive KL divergence:
\begin{equation*}
\mathrm{D}_{0}[p||q] = \mathrm{KL}[q||p] = \mathbb{E}_{q} \left[ \log \frac{q(\bo)}{ p(\bo) } \right];
\end{equation*}
2. Hellinger distance:
\begin{equation*}
\mathrm{D}_{0.5}[p||q] = 4\mathrm{Hel^2}[q||p] = 2 \int \left( \sqrt{p(\bo)} - \sqrt{q(\bo)} \right)^2 d\bo;
\end{equation*}
3. Inclusive KL divergence:
\begin{equation*}
\mathrm{D}_{1}[p||q] = \mathrm{KL}[p||q] = \mathbb{E}_{p} \left[ \log \frac{p(\bo)}{ q(\bo) } \right].
\end{equation*}
Since $\alpha=0$ is used in VI and $\alpha = 1.0$ is used in EP, in later sections we will also refer to these alpha settings as the VI value, Hellinger value, and EP value, respectively.
 
Power-EP, though providing a generic variational framework, does not scale with big data. It maintains approximating factors attached to every likelihood term $f_n(\bo)$, resulting in space complexity $\mathcal{O}(N)$ for the posterior approximation which is clearly undesirable. The recently proposed stochastic EP \cite{li2015sep} and BB-$\alpha$ \cite{hernandez2016black} inference methods reduce this memory overhead to $\mathcal{O}(1)$ by sharing these approximating factors. Moreover, optimisation in BB-$\alpha$ is done by descending the so called BB-$\alpha$ energy function, where Monte Carlo (MC) methods and automatic differentiation are also deployed to allow fast prototyping.

BB-$\alpha$ has been successfully applied to Bayesian neural networks for regression, classification \cite{hernandez2016black} and model-based reinforcement learning \cite{depeweg2016bnn_rl}. They all found that using $\alpha \neq 0$ often returns better approximations than the VI case. The reasons for the worse results of VI are two fold. From the perspective of inference, the zero-forcing behaviour of exclusive KL-divergences enforces the $q$ distribution to be zero in the region where the exact posterior has zero probability mass. Thus VI often fits to a local mode of the exact posterior and is over-confident in prediction. On hyper-parameter learning point of view, as the variational lower-bound is used as a (biased) approximation to the maximum likelihood objective, the learned model could be biased towards over-simplified cases \cite{turner2011two}. These problems could potentially be addressed by using $\alpha$-divergences. For example, inclusive KL encourages the coverage of the support set (referred as mass-covering), and when used in local divergence minimisation \cite{minka2005divergence}, it can fit an  approximation to a mode of $p(\bo)$ with better estimates of uncertainty. Moreover the BB-$\alpha$ energy provides a better approximation to the marginal likelihood as well, meaning that the learned model will be less biased and thus fitting the data distribution better \cite{li2016variational}. Hellinger distance seems to provide a good balance between zero-forcing and mass-covering, and empirically it has been found to achieve the best performance.

Given the success of $\alpha$-divergence methods, it is a natural idea to extend these algorithms to other classes of approximations such as dropout. 
However this task is non-trivial. First, the original formulation of BB-$\alpha$ energy is an ad hoc adaptation of power-EP energy (see appendix), which applies to exponential family $q$ distributions only.  Second, the energy function offers a limited intuitive interpretation to non-experts, thus of limited use for practitioners. Third and most importantly, a naive implementation of BB-$\alpha$ using dropout would bring in a prohibitive computational burden.
To see this, we first review the BB-$\alpha$ energy function in the general case \cite{li2016variational} given $\alpha \neq 0$:
\begin{equation}
\mathcal{L}_{\alpha}(q) = -\frac{1}{\alpha} \sum_n \log \mathbb{E}_q \left[ \left(  \frac{f_n(\bo) p_0(\bo)^{\frac{1}{N}}}{q(\bo)^{\frac{1}{N}}} \right)^{\alpha} \right].
\label{eq:bbalpha_original}
\end{equation}
One could verify that this is the same energy function as presented in \cite{hernandez2016black} by considering $q$ an exponential family distribution. In practice (\ref{eq:bbalpha_original}) might be intractable, hence an MC approximation is introduced:
\begin{equation}
\mathcal{L}_{\alpha}^{\text{MC}}(q) = -\frac{1}{\alpha} \sum_n \log \frac{1}{K} \sum_k \left[ \left(  \frac{f_n(\boh_k) p_0(\boh_k)^{\frac{1}{N}}}{q(\boh_k)^{\frac{1}{N}}} \right)^{\alpha} \right]
\label{eq:bbalpha_original_mc}
\end{equation}
with $\quad \boh_k \sim q(\bo)$. This is a biased approximation as the expectation in (\ref{eq:bbalpha_original}) is computed before taking the logarithm. But empirically \citet{hernandez2016black} showed that the bias introduced by the MC approximation is often dominated by the variance of the samples, meaning that the effect of the bias is negligible.
When $\alpha \rightarrow 0$ it returns the \emph{variational free energy} (the VI objective)
\begin{equation}
\mathcal{L}_{0}(q) = \mathcal{L}_{\text{VFE}}(q) = \mathrm{KL}[q||p_0] - \sum_n \mathbb{E}_{q} \left[ \log f_n(\bo) \right],
\label{eq:vfe}
\end{equation}
and the corresponding MC approximation $\mathcal{L}_{\text{VFE}}^{\text{MC}}$ becomes an unbiased estimator of $\mathcal{L}_{\text{VFE}}$. Also $\mathcal{L}_{\alpha}^{\text{MC}} \rightarrow \mathcal{L}_{\text{VFE}}^{\text{MC}}$ as the number of samples $K \rightarrow 1$.

The original paper \cite{hernandez2016black} proposed a naive implementation which directly evaluates the MC estimation (\ref{eq:bbalpha_original_mc}) with samples $\boh_k \sim q(\bo)$.
However as discussed before, dropout implicitly samples different masked weight matrices $\boh \sim q$ for different data points. This indicates that the naive approach, when applied to dropout approximation, would gather all these samples for all $M$ datapoints in a mini-batch (i.e.~$MK$ sets of neural network weight matrices in total), which brings prohibitive cost if the network is wide and deep. 
Interestingly, the minimisation of the variational free energy ($\alpha = 0$) with the dropout approximation can be computed very efficiently. The main reason for this success is due to the additive structure of the variational free energy: no evaluation of $q$ density is required if the ``regulariser'' $\mathrm{KL}[q||p_0]$ can be computed/approximated efficiently. In the following section we propose an improved version of BB-$\alpha$ energy to allow applications with dropout and other flexible approximation structures.

\section{A New Reparameterisation of BB-$\alpha$ Energy}

We propose a reparamterisation of the BB-$\alpha$ energy to reduce the computational overhead, which uses the so called ``cavity distributions''.
First we denote $\tilde{q}(\bo)$ as a free-form cavity distribution, and write the approximate posterior $q$ as
\begin{equation}
q(\bo) = \frac{1}{Z_q} \tilde{q}(\bo) \left( \frac{\tilde{q}(\bo)}{p_0(\bo)} \right)^{\frac{\alpha}{N - \alpha}},
\label{eq:cavity}
\end{equation}
where we assume $Z_q < +\infty$ is the normalising constant to ensure $q$ a valid distribution. When $\alpha/N \rightarrow 0$, the unnormalised density in (\ref{eq:cavity}) converges to $\tilde{q}(\bo)$ for every $\bo$, and $Z_q \rightarrow 1$ by the assumption of $Z_q < +\infty$ \cite{van_erven2014renyi}. Hence $q \rightarrow \tilde{q}$ when $\alpha/N \rightarrow 0$, and this happens for example when we choose $\alpha \rightarrow 0$, or $N \rightarrow +\infty$ as well as when $\alpha$ grows sub-linearly to $N$. Now we rewrite the BB-alpha energy in terms of $\tilde{q}$:
\begin{equation*}
\begin{aligned}
\mathcal{L}_{\alpha}( q ) &= -\frac{1}{\alpha} \sum_n \log \int \left( \frac{1}{Z_q} \tilde{q}(\bo) \left( \frac{\tilde{q}(\bo)}{p_0(\bo)} \right)^{\frac{\alpha}{N - \alpha}} \right)^{1 - \frac{\alpha}{N}} \\ 
&\quad\quad\quad\quad\quad\quad\quad\quad p_0(\bo)^{\frac{\alpha}{N}} f_n(\bo)^{\alpha} d \bo \\
&= \frac{N}{\alpha} (1 - \frac{\alpha}{N}) \log \int \tilde{q}(\bo) \left( \frac{\tilde{q}(\bo)}{p_0(\bo)} \right)^{\frac{\alpha}{N - \alpha}} d \bo \\
&\quad\quad\quad\quad\quad\quad - \frac{1}{\alpha} \sum_n \log \mathbb{E}_{\tilde{q}} \left[ f_n(\bo)^{\alpha} \right] \\
&= \mathrm{R}_{\beta}[\tilde{q}||p_0] - \frac{1}{\alpha} \sum_n \log \mathbb{E}_{\tilde{q}} \left[ f_n(\bo)^{\alpha} \right], ~\beta = \frac{N}{N - \alpha},
\end{aligned}
\end{equation*} 
where $\mathrm{R}_{\beta}[\tilde{q}||p_0]$ represents the \emph{R{\'e}nyi divergence} (\citet{renyi1961divergence}, discussed in the appendix) of order $\beta$. We note again that when $\frac{\alpha}{N} \rightarrow 0$ the new energy $\mathcal{L}_{\alpha}(\tilde{q})$ converges to $\mathcal{L}_{\text{VFE}}(\tilde{q})$ as well as $q \rightarrow \tilde{q}$. More importantly, $\mathrm{R}_{\beta}[\tilde{q}||p_0] \rightarrow \mathrm{KL}[\tilde{q}||p_0] = \mathrm{KL}[q||p_0]$ provided $\mathrm{R}_{\beta}[\tilde{q}||p_0] < +\infty$ (which holds when assuming $Z_q < +\infty$) and $\frac{\alpha}{N} \rightarrow 0$. 

This means that for a constant $\alpha$ that scales sub-linearly with $N$, in large data settings we can further approximate the BB-$\alpha$ energy as 
\begin{equation*}
\mathcal{L}_{\alpha}( q ) \approx \tilde{\mathcal{L}}_{\alpha}(q) = \mathrm{KL}[q||p_0] - \frac{1}{\alpha} \sum_n \log \mathbb{E}_{q} \left[ f_n(\bo)^{\alpha} \right].
\end{equation*}
Note that here we also use the fact that now $q \approx \tilde{q}$. Critically, the proposed reparameterisation is continuous in $\alpha$, and by taking $\alpha \rightarrow 0$ the variational free-energy (\ref{eq:vfe}) is again recovered.

Given a loss function $l(\cdot, \cdot)$, e.g.~$l_2$ loss in regression or cross entropy in classification, we can define the (unnormalised) likelihood term $f_n(\bo) \propto p(\y_n|\bm{x}_n, \bo) \propto \exp [-l(\y_n, \f^{\bo}(\bm{x}_n))] $, e.g.~see \cite{lecun2006energy}\footnote{We note that $f_n(\bo)$ does not need to be a normalised density of $y_n$ unless one would like to optimise the hyper parameters associated with $f_n$.}.
Swapping $f_n(\bo)$ for this last expression, and approximating
the expectation over $q$ using Monte Carlo sampling, we obtain our proposed minimisation objective: 
\begin{align}
\label{eq:bbalpha_new}
\tilde{\mathcal{L}}^{\text{MC}}_{\alpha}(q) &= \mathrm{KL}[q||p_0] + \text{const} \\
&\quad - \frac{1}{\alpha} \sum_n \logsumexp [-\alpha l(y_n, \f^{\boh_k}(\bm{x}_n))] 
\notag
\end{align}
with $\logsumexp$ being the log-sum-exp operator over $K$ samples from the approximate posterior $\boh_k \sim q(\bo)$.
This objective function also approximates the marginal likelihood.
Therefore, compared to the original formulation (\ref{eq:bbalpha_original}), the improved version (\ref{eq:bbalpha_new}) is considerably simpler (both to implement and to understand), has a similar form to standard objective functions used in deep learning research, yet remains an approximate Bayesian inference algorithm.

To gain some intuitive understanding of this objective, we observe what it reduces to for different $\alpha$ and $K$ settings. By selecting $\alpha=1$ the per-point predictive log-likelihood $\log \mathbb{E}_q[p(y_n|\x_n, \bo)]$ is directly optimised. On the other hand, picking the VI value ($\alpha \rightarrow 0$) would focus on increasing the training accuracy $\mathbb{E}_q [\log p(y_n|\x_n, \bo)]$. The Hellinger value could be used to achieve a balance between reducing training error and improving predictive likelihood, which has been found to be desirable \cite{hernandez2016black,depeweg2016bnn_rl}. Lastly, for $K=1$ the log-sum-exp disappears, the $\alpha$'s cancel out, and the original (stochastic) VI objective is recovered.

In summary, our proposal modifies the loss function by multiplying it by $\alpha$ and then performing log-sum-exp with a sum over multiple stochastic forward passes sampled from the BNN approximate posterior. 
The remaining KL-divergence term (between $q$ and the prior $p$) can often be approximated. It can be viewed as a regulariser added to the objective function, and reduces to $L_2$-norm regulariser for certain popular $q$ choices \citep{Gal2016Uncertainty}.

\subsection{Dropout BB-$\alpha$}
We now provide a concrete example where the approximate distribution is defined by dropout.  With dropout VI, MC samples are used to approximate the expectation w.r.t.\ $q$, which in practice is implemented as performing \textit{stochastic forward passes} through the dropout network -- i.e.\ given an input $\x$, the input is fed through the network and a new dropout mask is sampled and applied at each dropout layer. This gives a stochastic output -- a sample from the dropout network on the input $\x$. A similar approximation is used in our case as well, where to implement the MC sampling in eq.\ \eqref{eq:bbalpha_new} we perform multiple stochastic forward passes through the network.

Recall the neural network $\f^\bo(\x)$ is parameterised by the variable $\bo$. In classification, cross entropy is often used as the loss function
\begin{gather}
\sum_n l(\y_n, \mathbf{p}^\bo(\x_n)) = \sum_n -\y_n^T \log \mathbf{p}^\bo(\x_n),
\\
\mathbf{p}^\bo(\x_n) = \softmax (
\f^{\bo}(\bm{x}_n)
), \notag
\end{gather}
where the label $\y_n$ is a one-hot binary vector, and the network output $\softmax (\f^\bo(\x_n))$ encodes the probability vector of class assignments.
Applying the re-formulated BB-$\alpha$ energy (\ref{eq:bbalpha_new}) with a Bayesian equivalent of the network, we arrive at the objective function
\begin{align}
\tilde{\mathcal{L}}^{\text{MC}}_{\alpha}(q) &=
%
%
\sum_{i} p_i ||\M_i||_2^2 - \frac{1}{\alpha} \sum_n \y_n^T \log \frac{1}{K} \sum_k (
\mathbf{p}^{\boh_k}(\x_n)
)^{\alpha}
\notag \\
&= \frac{1}{\alpha} \sum_n l \left( \y_n, 
\frac{1}{K} \sum_k 
\mathbf{p}^{\boh_k}(\x_n)
^{\alpha}
\right) + \sum_i L_2
(\M_i)
\end{align}
%
with $\{ \mathbf{p}^{\boh_k}(\x_n) \}_{k=1}^K$ being $K$ stochastic network outputs on input $\x_n$,
$p_i$ equals to one minus the dropout rate of the $i$th layer, and the $L_2$ regularization terms coming from an approximation to the KL-divergence \citep{Gal2016Uncertainty}. I.e.\ we raise network probability outputs to the power $\alpha$ and average them as an input to the standard cross entropy loss.
Taking $\alpha \neq 1$ can be viewed as training the neural network with  an adjusted ``power'' loss, regularized by an $L_2$ norm. 
Implementing this induced loss with Keras \citep{keras2015} is as simple as a few lines of Python. A code snippet is given in Figure \ref{code}, with more details in the appendix.

In regression problems, the loss function is defined as $l(\bm{y}, \f^{\bo}(\bm{x})) = \frac{\tau}{2} ||\bm{y} - \f^{\bo}(\bm{x})||_2^2$ and the likelihood term can be interpreted as $\y \sim \mathcal{N}(\bm{y}; \f^{\bo}(\bm{x}), \tau^{-1} \bm{I})$. Plugging this into the energy function returns the following objective
\begin{align}
\tilde{\mathcal{L}}^{\text{MC}}_{\alpha}(q) &= - \frac{1}{\alpha} \sum_n \logsumexp \left[-\frac{\alpha \tau}{2} ||\y_n - \f^{\boh_k}(\bm{x}_n)||_2^2 \right] \notag \\ 
&\qquad + \frac{ND}{2} \log \tau + \sum_{ i} p_i ||\bm{M}_i||_2^2,
\end{align}
%
with $\{ \f^{\boh_k}(\x_n) \}_{k=1}^K$ being $K$ stochastic forward passes on input $\x_n$.
Again, this is reminiscent of the $l_2$ objective in standard deep learning, and can be implemented by simply passing the input through the dropout network multiple times, collecting the stochastic outputs, and feeding the set of outputs through our new BB-alpha loss function.

\begin{figure}[t!]
\small
\begin{minted}{python}
def softmax_cross_ent_with_mc_logits(alpha):
  def loss(y_true, mc_logits):
    # mc_logits: MC samples of shape MxKxD
    mc_log_softmax = mc_logits \
      - K.max(mc_logits, axis=2, keepdims=True)
    mc_log_softmax = mc_log_softmax - \ 
      logsumexp(mc_log_softmax, 2)
    mc_ll = K.sum(y_true*mc_log_softmax,-1)
    return -1./alpha * (logsumexp(alpha * \
      mc_ll, 1) + K.log(1.0 / K_mc))
  return loss
\end{minted}
\vspace{-4mm}
\caption{Code snippet for our induced classification loss.}
\label{code}
\end{figure}

\section{Experiments}
We test the reparameterised BB-$\alpha$ on Bayesian NNs with the dropout approximation. We assess the proposed inference in regression and classification tasks on standard benchmarking datasets, comparing different values of $\alpha$. We further assess the training time trade-off between our technique and VI, and study the properties of our model's uncertainty on out-of-distribution data points. This last experiment leads us to propose a technique that could be used to identify adversarial image attacks.

\subsection{Regression}

The first experiment considers Bayesian neural network regression with approximate posterior induced by dropout. We use benchmark UCI datasets\footnote{\url{http://archive.ics.uci.edu/ml/datasets.html}} that have been tested in related literature. 
The model is a single-layer neural network with 50 ReLU units for all datasets except for Protein and Year, which use 100 units. We consider $\alpha \in \{0.0, 0.5, 1.0\}$ in order to examine the effect of mass-covering/zero-forcing behaviour in dropout. MC approximation with $K=10$ samples is also deployed to compute the energy function. Other initialisation settings are largely taken from \cite{li2016variational}.

\begin{figure*}[t]
\centering
\begin{minipage}{0.47\linewidth}
\includegraphics[width=1.0\linewidth]{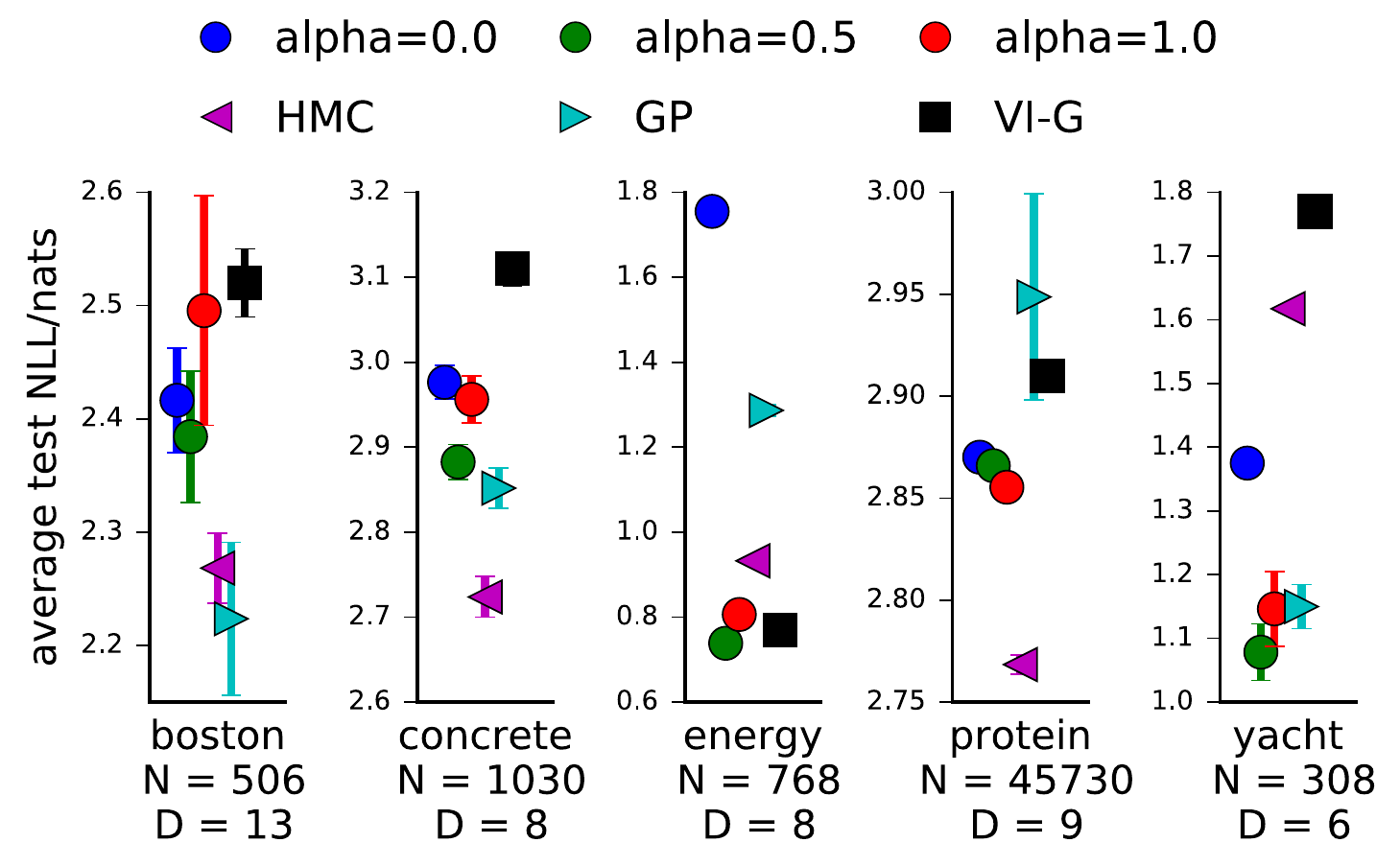}
\caption{Negative test-LL results for Bayesian NN regression. The lower the better. Best viewed in colour.}
\label{fig:bnn_regression_ll}
\end{minipage}
\hspace{3mm}
\begin{minipage}{0.47\linewidth}
\centering
\includegraphics[width=1.0\linewidth]{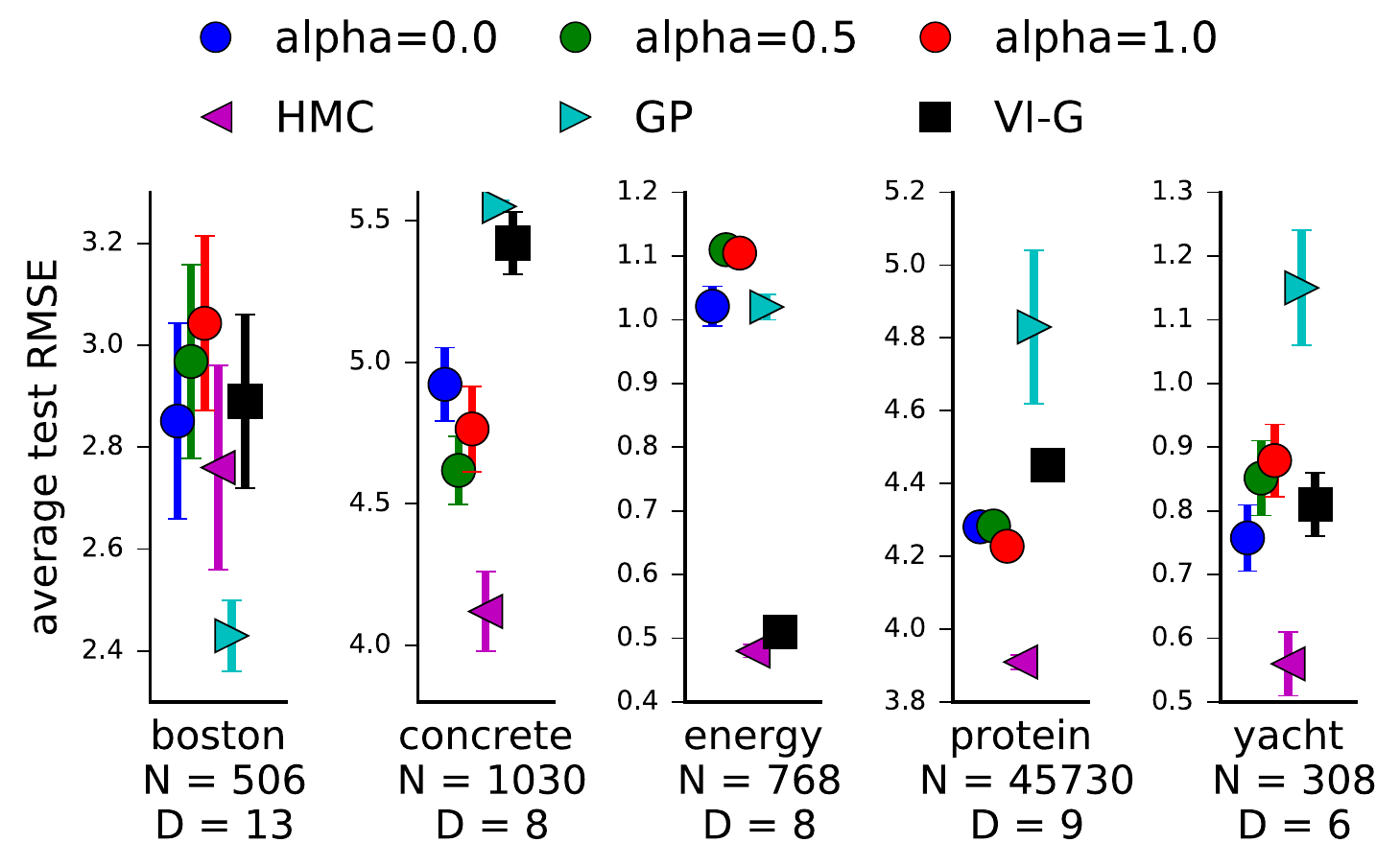}
\caption{Test RMSE results for Bayesian NN regression. The lower the better. Best viewed in colour.}
\label{fig:bnn_regression_rmse}
\end{minipage}
\end{figure*}

We summarise the test negative log-likelihood (LL) and RMSE with standard error (across different random splits) for selected datasets in Figure \ref{fig:bnn_regression_ll} and \ref{fig:bnn_regression_rmse}, respectively. The full results are provided in the appendix. 
Although optimal $\alpha$ may vary for different datasets, using non-VI values has significantly improved the test-LL performances, while remaining comparable in test error metric.
In particular, $\alpha = 0.5$ produced overall good results for both test LL and RMSE, which is consistent with previous findings.
As a comparison we also include test performances of a BNN with a Gaussian approximation (VI-G) \cite{li2016variational}, a BNN with HMC, and a sparse Gaussian process model with 50 inducing points \cite{bui2016dgp}. In test-LL metric our best dropout model out-performs the Gaussian approximation method on almost all datasets, and for some datasets is on par with HMC which is the current gold standard for Bayesian neural works, and with the GP model that is known to be superior in regression.


\subsection{Classification}

\begin{figure}[b!]
\center
\begin{subfigure}[t]{0.49\linewidth}
\includegraphics[width=\linewidth]{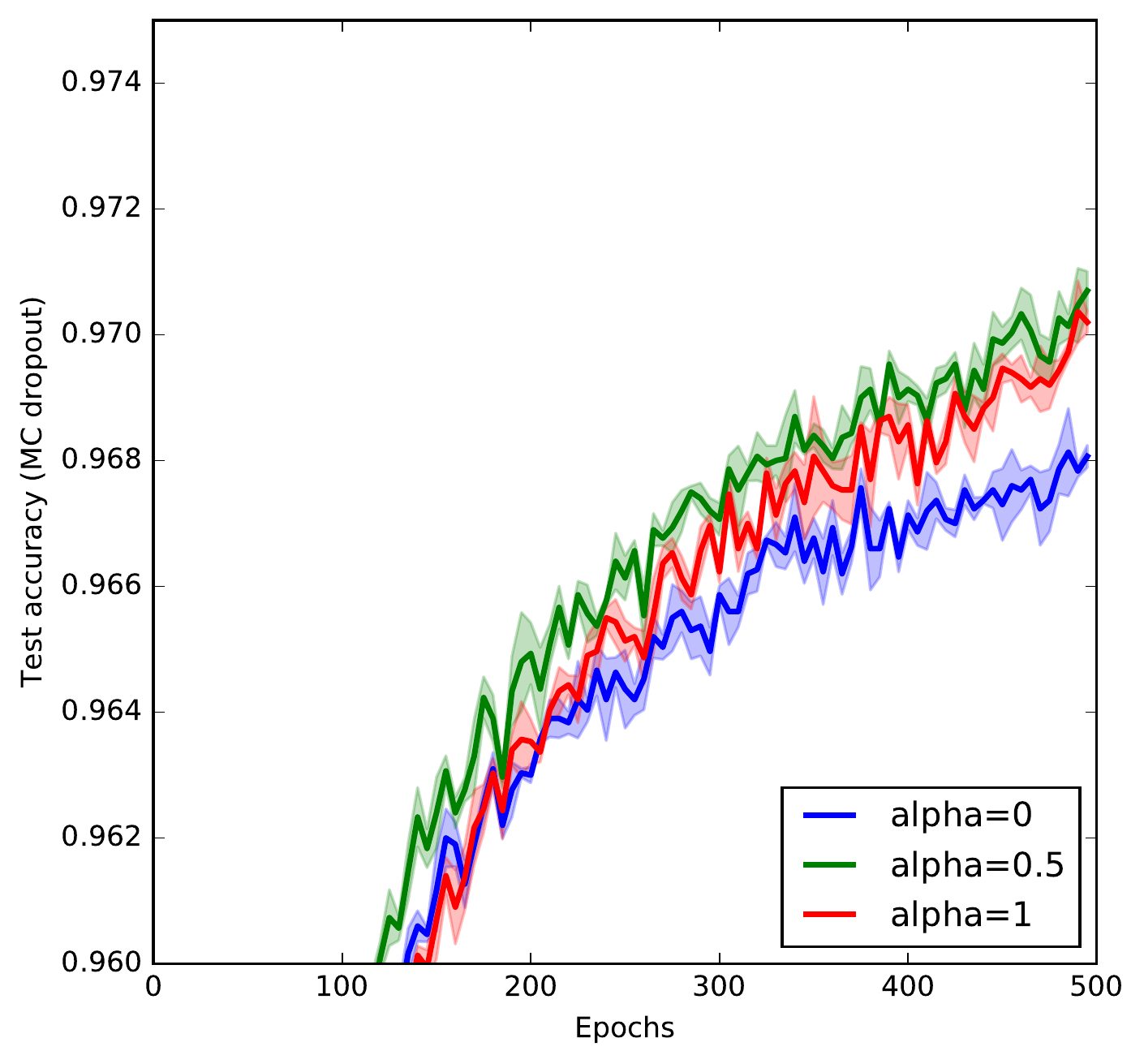}
\caption{Fully connected NN test accuracy}
\end{subfigure}
\begin{subfigure}[t]{0.49\linewidth}
\includegraphics[width=\linewidth]{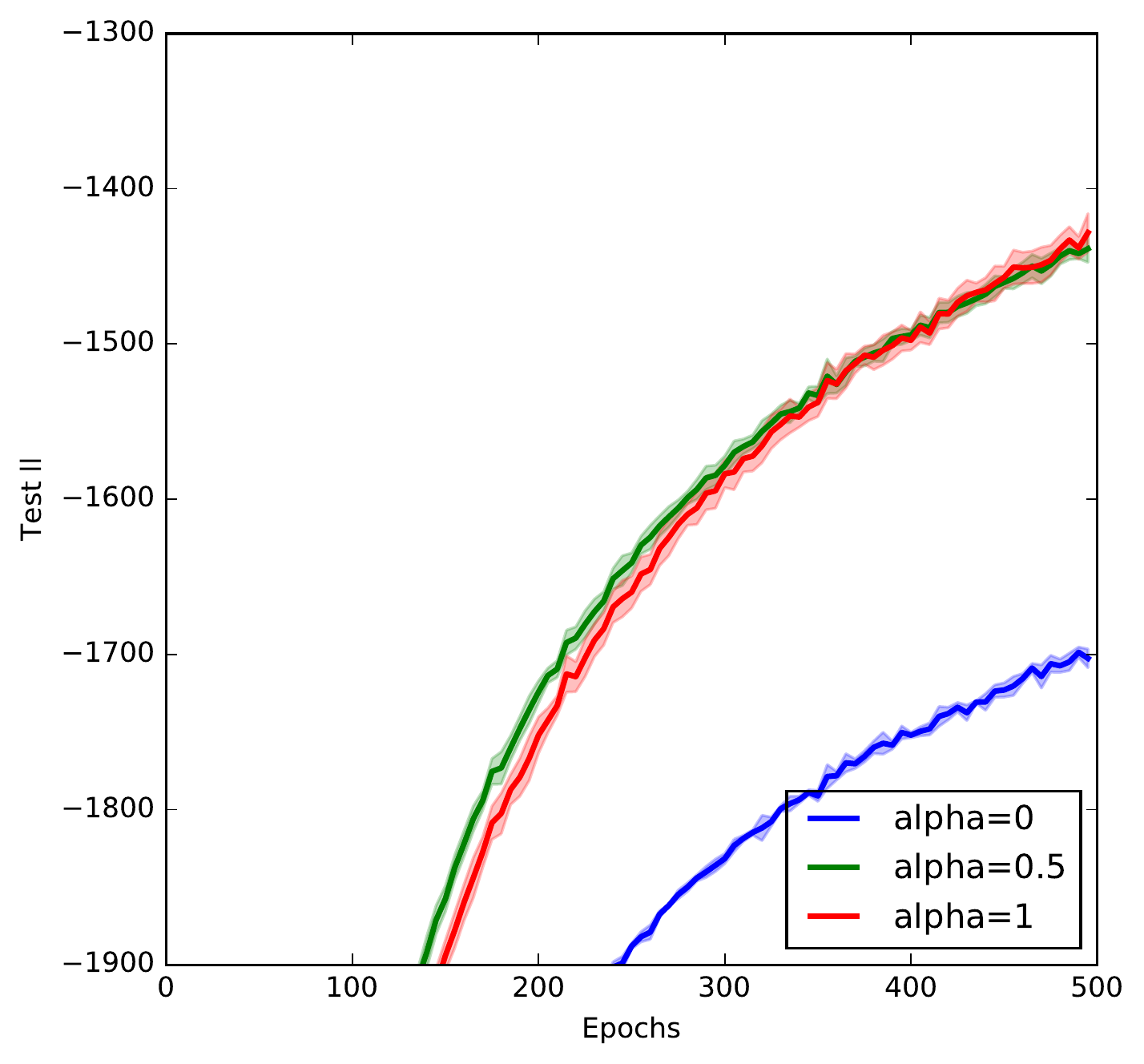}
\caption{Fully connected NN test log likelihood}
\end{subfigure}
\vspace{2mm}
\caption{MNIST test accuracy and test log likelihood for a fully connected NN in a classification task.}
\label{fig:clas1}
\end{figure}

We further experiment with a classification task, comparing the accuracy of the various $\alpha$ values on the MNIST benchmark \citep{lecun1998mnist}. We assessed a fully connect NN with 2 hidden layers and 100 units in each layer. We used dropout probability 0.5 and $\alpha \in \{0, 0.5, 1 \}$. Again, we use $K=10$ samples at training time for all $\alpha$ values, and $K_\text{test}=100$ samples at test time. We use weight decay $10^{-6}$, which is equivalent to prior lengthscale $l^2=0.1$ \citep{gal2016dropout}. We repeat each experiment three times and plot mean and standard error. Test RMSE as well as test log likelihood are given in Figure \ref{fig:clas1}. As can be seen, Hellinger value $\alpha=0.5$ gives best test RMSE, with test log likelihood matching that of the EP value $\alpha=1$. The VI value $\alpha=0$ under-performs according to both metrics.


\begin{figure}[b!]
\center
\begin{subfigure}[t]{0.49\linewidth}
\includegraphics[width=\linewidth]{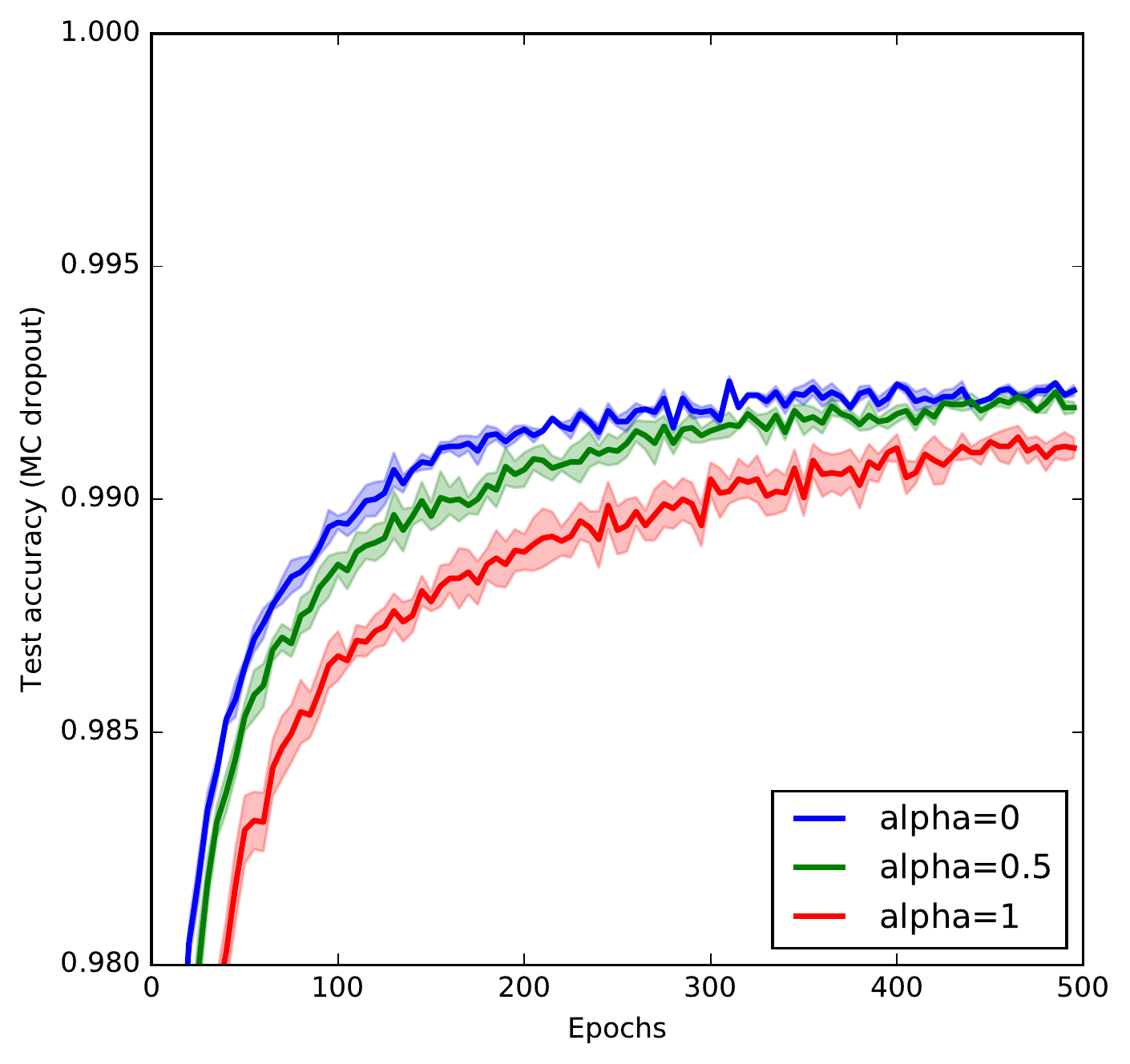}
\caption{CNN test accuracy}
\end{subfigure}
\begin{subfigure}[t]{0.49\linewidth}
\includegraphics[width=\linewidth]{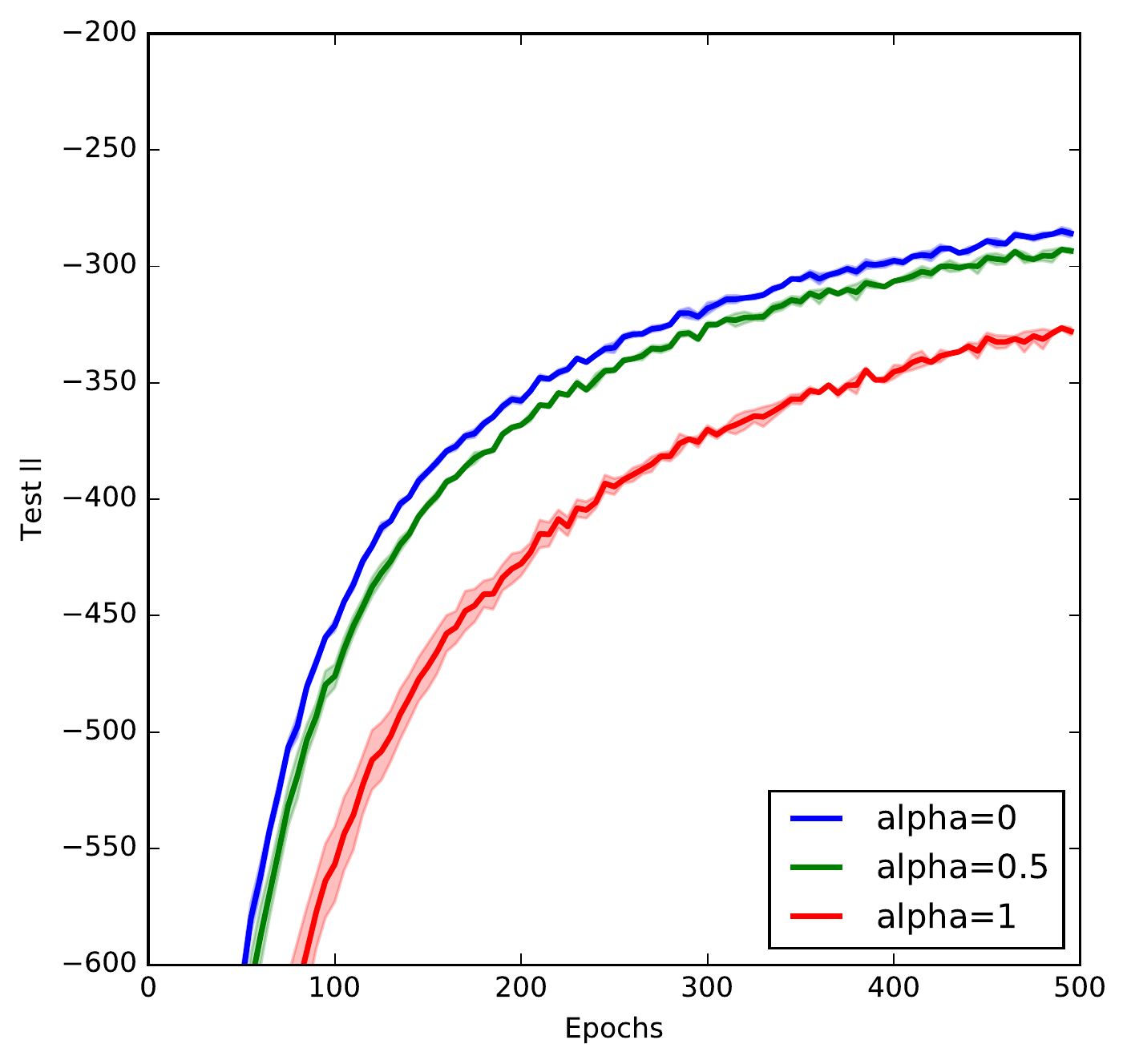}
\caption{CNN test log likelihood}
\end{subfigure}
\vspace{2mm}
\caption{MNIST test accuracy and test log likelihood for a convolutional neural network in a classification task.}
\label{fig:clas2}
\end{figure}

We next assess a convolutional neural network model (CNN). For this experiment we use the standard CNN example given in \citep{keras2015} with 32 convolution filters, 100 hidden units at the top layer, and dropout probability $0.5$ before each fully-connected layer. Other settings are as before. Average test accuracy and test log likelihood are given in Figure \ref{fig:clas2}.
In this case, VI value $\alpha=0$ seems to supersede the EP value $\alpha=1$, and performs similarly to the Hellinger value $\alpha=0.5$ according to both metrics.

\subsection{Detecting Adversarial Examples}

\begin{figure*}[t]
\center
\begin{minipage}{0.48\linewidth}
\includegraphics[width=1\linewidth]{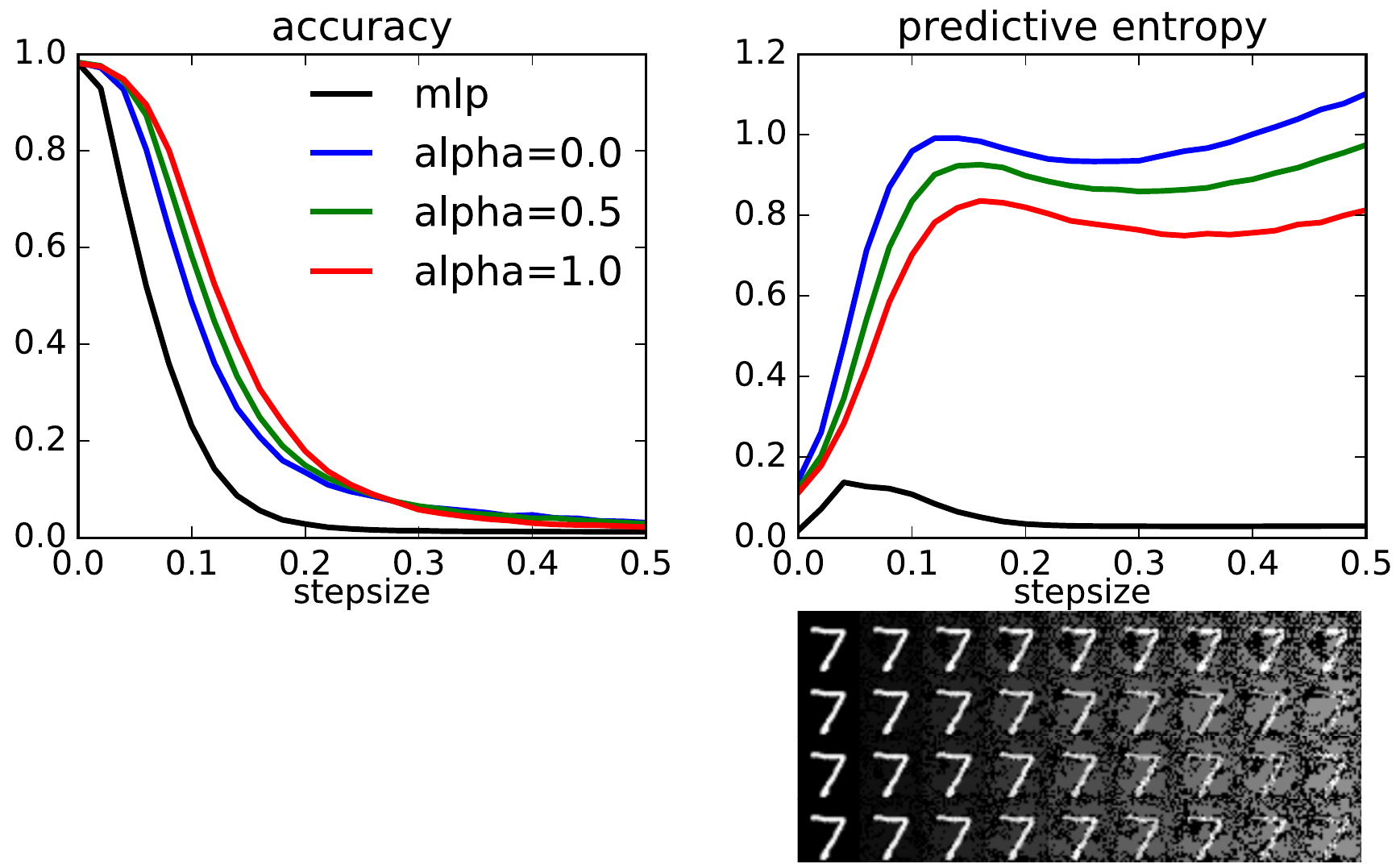}
\caption{Un-targeted attack: classification accuracy results as a function of perturbation stepsize. The adversarial examples are shown for (from top to bottom) NN and BNN trained with dropout and $\alpha = 0.0, 0.5, 1.0$.}
\label{fig:attack_untargeted}
\end{minipage}
\hspace{3mm}
\begin{minipage}{0.48\linewidth}
\includegraphics[width=1\linewidth]{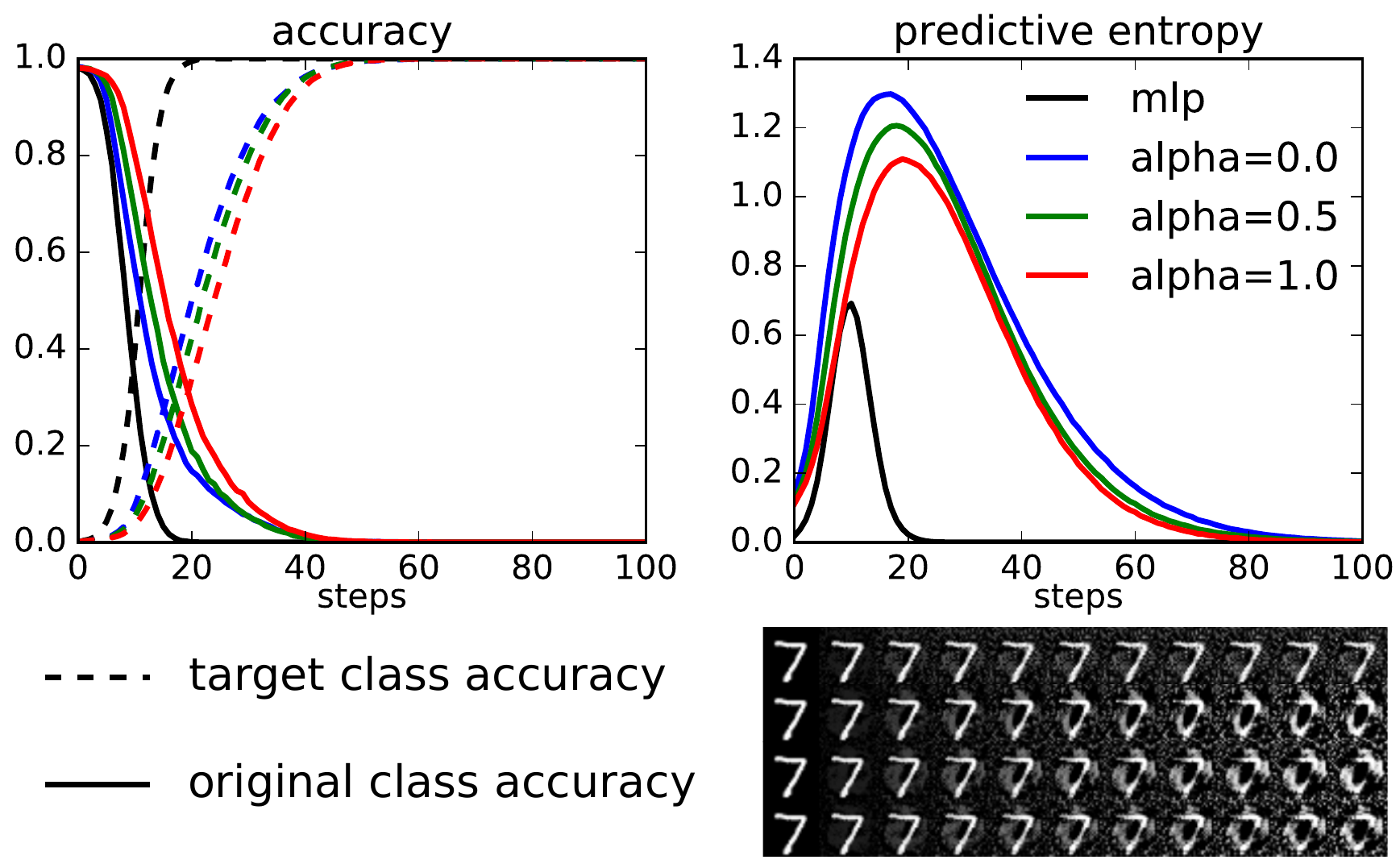}
\caption{Targeted attack: classification accuracy results as a function of the number of iterative gradient steps. The adversarial examples are shown for (from top to bottom) NN and BNN trained with dropout and $\alpha = 0.0, 0.5, 1.0$.}
\label{fig:attack_targeted}
\end{minipage}
\end{figure*}

The third set of experiments considers adversarial attacks on dropout trained Bayesian neural networks. 
Bayesian neural networks' uncertainty increases on examples far from the data distribution. We test the hypothesis that certain techniques for generating adversarial examples will give images that lie outside of the image manifold, i.e.\ far from the data distribution (note though that there exist techniques that will guarantee the images staying near the data manifold, by minimising the perturbation used to construct the adversarial example).
By assessing our BNN uncertainty, we should see increased uncertainty for adversarial images if they indeed lie outside of the training data distribution.
The tested model is a fully connected network with 3 hidden layers of 1000 units. The dropout trained models are also compared to a benchmark NN with the same architecture but trained by maximum likelihood. The adversarial examples are generated on MNIST test data that is normalised to be in the range $[0, 1]$. For the dropout trained networks we perform MC dropout prediction at test time with $K_{\text{test}} = 10$ MC samples.

The first attack in consideration is the Fast Gradient Sign (FGS) method \cite{goodfellow2014explaining}. This is an un-targeted attack, which attempts to reduces the maximum value of the predicted class label probability 
\begin{equation*}
\x_{\text{adv}} = \x - \eta \cdot \text{sgn}(\nabla_{\x} \max_{y} \log p(y|\x)).
\end{equation*}
We use the single gradient step FGS implemented in Cleverhans \cite{papernot2016cleverhans} with the stepsize $\eta$ varied between 0.0 and 0.5. The left panel in Figure \ref{fig:attack_untargeted} demonstrates the classification accuracy on adversarial examples, which shows that the dropout networks, especially the one trained with $\alpha=1.0$, are significantly more robust to adversarial attacks compared to the deterministic NN. More interestingly, the test data examples and adversarial images can be told-apart by investigating the uncertainty representation of the dropout models. In the right panel of Figure \ref{fig:attack_untargeted} we depict the predictive entropy computed on the neural network output probability vector, and show example corresponding adversarial images below the axis for each corresponding stepsize. Clearly the deterministic NN model produces over-confident predictions on adversarial samples, e.g.~it predicts the wrong label very confidently even when the input is still visually close to digit ``7'' ($\eta = 0.2$). While dropout models, though producing wrong labels, are very uncertain about their predictions. This uncertainty keeps increasing as we move away from the data manifold. Hence the dropout networks are much more immunised from noise-corrupted inputs, as they can be detected using uncertainty estimates in this example.

The second attack we consider is a targeted version of FSG \cite{carlini2016towards}, which maximises the predictive probability of a selected class instead. As an example, we fix class 0 as the target and apply the iterative gradient-base attack to all non-zero digits in test data. At step $t$, the adversarial output is computed as
\begin{equation*}
\x_{\text{adv}}^t = \x_{\text{adv}}^{t-1} + \eta \cdot \text{sgn}(\nabla_{\x} \log p(y_{\text{target}} | \x_{\text{adv}}^{t-1} ) ),
\end{equation*}
where the stepsize $\eta$ is fixed at $0.01$ in this case.
Results are presented in the left panel of Figure \ref{fig:attack_targeted}, and again dropout trained models are more robust to this attack compared with the deterministically trained NN. Similarly these adversarial examples could be detected by the Bayesian neural networks' uncertainty, by examining the predictive entropy. By visually inspecting the generated adversarial examples in the right panel of Figure \ref{fig:attack_targeted}, it is clear that the NN overconfidently classifies a digit 7 to class 0. On the other hand, the dropout models are still fairly uncertain about their predictions even after 40 gradient steps. More interestingly, running this iterative attack on dropout models produces a smooth interpolation between different digits, and when the model is confident on predicting the target class, the corresponding adversarial images are visually close to digit zero.

These initial results suggest that assessing the epistemic uncertainty of classification models can be used as a viable technique to identify adversarial examples. We would note though that we used this experiment to demonstrate our techniques' uncertainty estimates, and much more research is needed to solve the difficulties faced with adversarial inputs.

\subsection{Run time trade-off}

\begin{figure}[h]
\center
\begin{subfigure}[t]{0.49\linewidth}
\includegraphics[width=\linewidth]{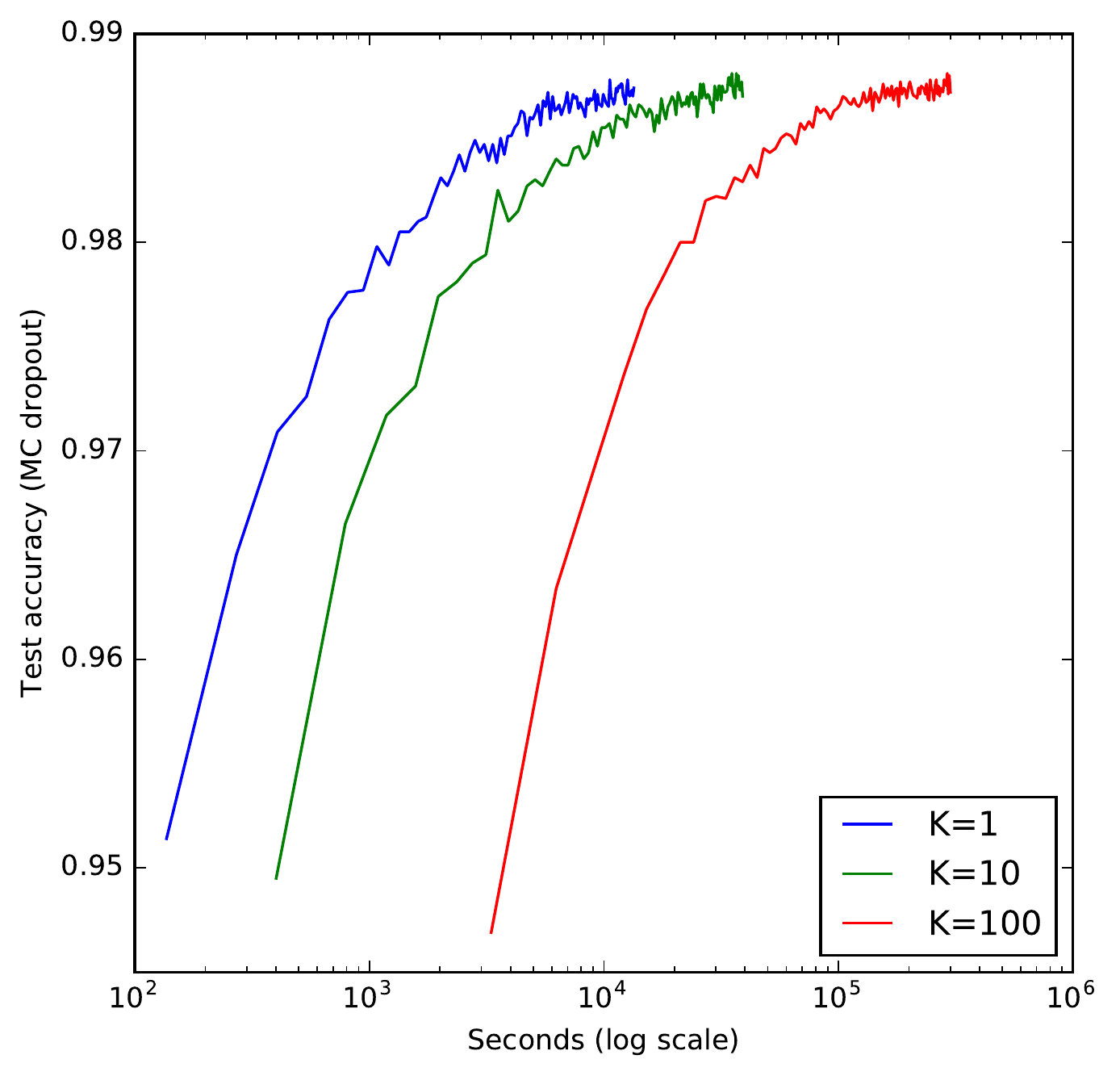}
\caption{Test accuracy}
\end{subfigure}
\begin{subfigure}[t]{0.49\linewidth}
\includegraphics[width=\linewidth]{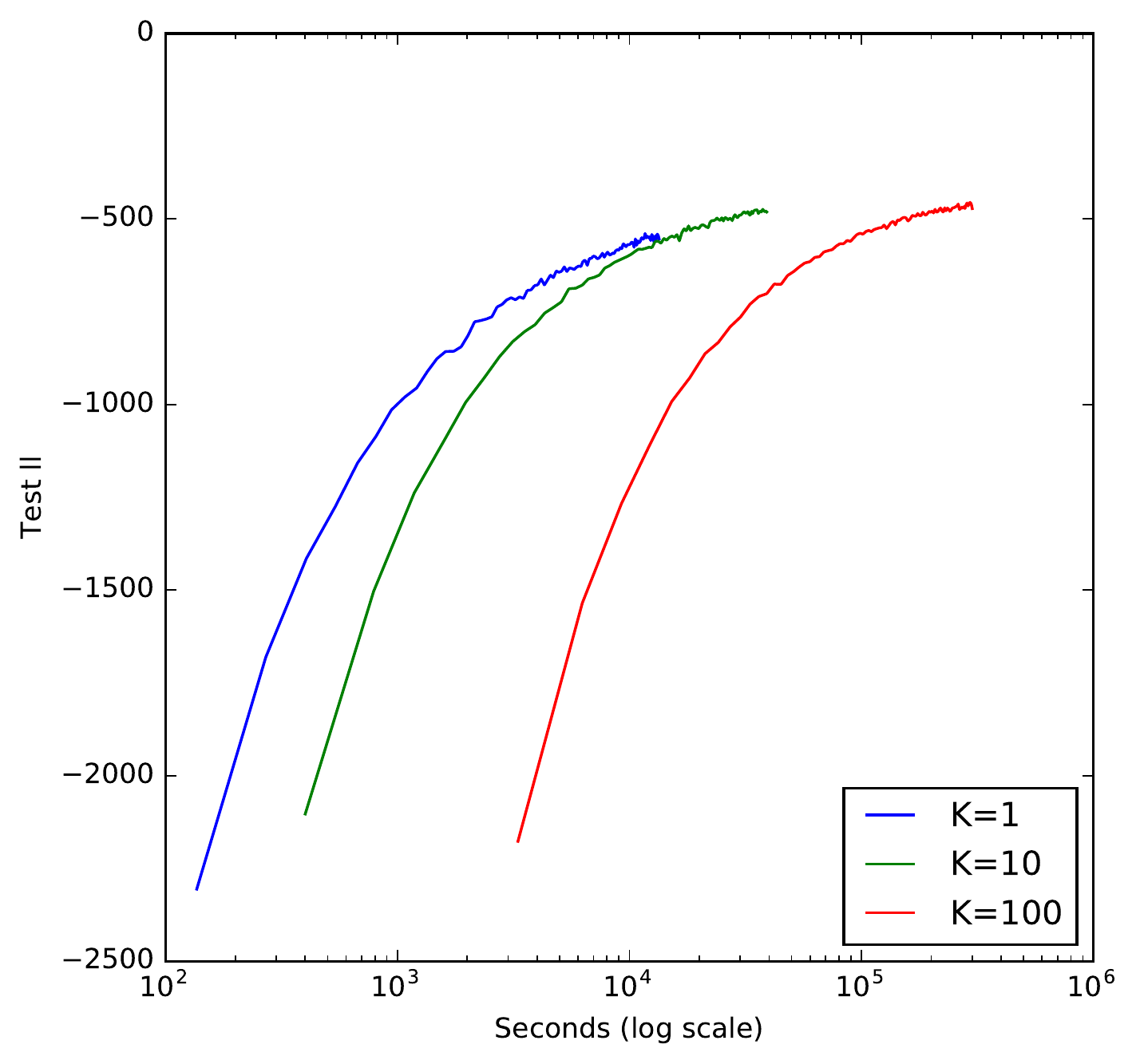}
\caption{Test log likelihood}
\end{subfigure}
\vspace{2mm}
\caption{Run time experiment on the MNIST dataset for different number of samples $K$.}
\label{fig:time}
\end{figure}

We finish the experiments section by assessing the running time trade-offs of using an increasing number of samples at training time. Unlike VI, in our inference we rely on a large number of samples to reduce estimator bias. When a small number of samples is used ($K=1$) our method collapses to standard VI. In Figure \ref{fig:time} we see both test accuracy as well as test log likelihood for a fully connected NN with four layers of 1024 units trained on the MNIST dataset, with $\alpha=1$. The two metrics are shown as a function of wall-clock run time for different values of $K \in \{ 1, 10, 100\}$. As can be seen, $K=1$ converges to test accuracy of $98.8\%$ faster than the other values of $K$, which converge to the same accuracy. On the other hand, when assessing test log likelihood, both $K=1$ and $K=10$ attain value $-600$ within 1000 seconds, but $K=10$ continues improving its test log likelihood and converges to value $-500$ after 3000 seconds. $K=100$ converges to the same value but requires much longer running time, possibly because of noise from other processes.

\section{Conclusions}
We presented a practical extension of the BB-alpha objective which allows us to use the technique with dropout approximating distributions. The technique often supersedes existing approximate inference techniques (even sparse Gaussian processes), and is easy to implement. A code snippet for our induced loss is given in the appendix. 

\section*{Acknowledgements}
YL thanks the Schlumberger Foundation
FFTF fellowship for supporting her PhD study.


{\small
\bibliography{references}

\begin{thebibliography}{50}
\providecommand{\natexlab}[1]{#1}
\providecommand{\url}[1]{\texttt{#1}}
\expandafter\ifx\csname urlstyle\endcsname\relax
  \providecommand{\doi}[1]{doi: #1}\else
  \providecommand{\doi}{doi: \begingroup \urlstyle{rm}\Url}\fi

\bibitem[Amari(1985)]{amari1985book}
Amari, Shun-ichi.
\newblock \emph{Differential-Geometrical Methods in Statistic}.
\newblock Springer, New York, 1985.

\bibitem[Amodei et~al.(2016)Amodei, Olah, Steinhardt, Christiano, Schulman, and
  Mane]{amodei2016concrete}
Amodei, Dario, Olah, Chris, Steinhardt, Jacob, Christiano, Paul, Schulman,
  John, and Mane, Dan.
\newblock Concrete problems in ai safety.
\newblock \emph{arXiv preprint arXiv:1606.06565}, 2016.

\bibitem[Angermueller \& Stegle(2015)Angermueller and
  Stegle]{angermueller2015multi}
Angermueller, C and Stegle, O.
\newblock Multi-task deep neural network to predict {CpG} methylation profiles
  from low-coverage sequencing data.
\newblock In \emph{NIPS MLCB workshop}, 2015.

\bibitem[Barber \& Bishop(1998)Barber and Bishop]{barber1998ensemble}
Barber, David and Bishop, Christopher~M.
\newblock Ensemble learning in {B}ayesian neural networks.
\newblock \emph{NATO ASI SERIES F COMPUTER AND SYSTEMS SCIENCES}, 168:\penalty0
  215--238, 1998.

\bibitem[Blundell et~al.(2015)Blundell, Cornebise, Kavukcuoglu, and
  Wierstra]{blundell2015weight}
Blundell, Charles, Cornebise, Julien, Kavukcuoglu, Koray, and Wierstra, Daan.
\newblock Weight uncertainty in neural network.
\newblock In \emph{ICML}, 2015.

\bibitem[Bui et~al.(2016)Bui, Hern{\'a}ndez-Lobato, Li, Hern{\'a}ndez-Lobato,
  and Turner]{bui2016dgp}
Bui, Thang~D, Hern{\'a}ndez-Lobato, Daniel, Li, Yingzhen, Hern{\'a}ndez-Lobato,
  Jos{\'e}~Miguel, and Turner, Richard~E.
\newblock Deep gaussian processes for regression using approximate expectation
  propagation.
\newblock In \emph{Proceedings of The 33rd International Conference on Machine
  Learning (ICML)}, 2016.

\bibitem[Carlini \& Wagner(2016)Carlini and Wagner]{carlini2016towards}
Carlini, Nicholas and Wagner, David.
\newblock Towards evaluating the robustness of neural networks.
\newblock \emph{arXiv preprint arXiv:1608.04644}, 2016.

\bibitem[Chollet(2015)]{keras2015}
Chollet, Francois.
\newblock Keras.
\newblock \url{https://github.com/fchollet/keras}, 2015.

\bibitem[Denker \& LeCun(1991)Denker and LeCun]{denker1991transforming}
Denker, John and LeCun, Yann.
\newblock Transforming neural-net output levels to probability distributions.
\newblock In \emph{Advances in Neural Information Processing Systems 3}.
  Citeseer, 1991.

\bibitem[Depeweg et~al.(2016)Depeweg, Hern{\'a}ndez-Lobato, Doshi-Velez, and
  Udluft]{depeweg2016bnn_rl}
Depeweg, Stefan, Hern{\'a}ndez-Lobato, Jos{\'e}~Miguel, Doshi-Velez, Finale,
  and Udluft, Steffen.
\newblock Learning and policy search in stochastic dynamical systems with
  bayesian neural networks.
\newblock \emph{arXiv preprint arXiv:1605.07127}, 2016.

\bibitem[Gal(2016)]{Gal2016Uncertainty}
Gal, Yarin.
\newblock \emph{Uncertainty in Deep Learning}.
\newblock PhD thesis, University of Cambridge, 2016.

\bibitem[Gal \& Ghahramani(2016{\natexlab{a}})Gal and
  Ghahramani]{Gal2016Bayesian}
Gal, Yarin and Ghahramani, Zoubin.
\newblock Bayesian convolutional neural networks with {B}ernoulli approximate
  variational inference.
\newblock \emph{ICLR workshop track}, 2016{\natexlab{a}}.

\bibitem[Gal \& Ghahramani(2016{\natexlab{b}})Gal and
  Ghahramani]{gal2016dropout}
Gal, Yarin and Ghahramani, Zoubin.
\newblock Dropout as a {B}ayesian approximation: Representing model uncertainty
  in deep learning.
\newblock \emph{ICML}, 2016{\natexlab{b}}.

\bibitem[Goodfellow et~al.(2014)Goodfellow, Shlens, and
  Szegedy]{goodfellow2014explaining}
Goodfellow, Ian~J, Shlens, Jonathon, and Szegedy, Christian.
\newblock Explaining and harnessing adversarial examples.
\newblock \emph{arXiv preprint arXiv:1412.6572}, 2014.

\bibitem[Graves(2011)]{graves2011practical}
Graves, Alex.
\newblock Practical variational inference for neural networks.
\newblock In \emph{Advances in Neural Information Processing Systems}, pp.\
  2348--2356, 2011.

\bibitem[Hellinger(1909)]{hellinger1909neue}
Hellinger, Ernst.
\newblock Neue begr{\"u}ndung der theorie quadratischer formen von
  unendlichvielen ver{\"a}nderlichen.
\newblock \emph{Journal f{\"u}r die reine und angewandte Mathematik},
  136:\penalty0 210--271, 1909.

\bibitem[Hernandez-Lobato \& Adams(2015)Hernandez-Lobato and
  Adams]{hernandez2015probabilistic}
Hernandez-Lobato, Jose~Miguel and Adams, Ryan.
\newblock Probabilistic backpropagation for scalable learning of {B}ayesian
  neural networks.
\newblock In \emph{ICML}, 2015.

\bibitem[Hern{\'a}ndez-Lobato et~al.(2016)Hern{\'a}ndez-Lobato, Li,
  Hern{\'a}ndez-Lobato, Bui, and Turner]{hernandez2016black}
Hern{\'a}ndez-Lobato, Jos{\'e}~Miguel, Li, Yingzhen, Hern{\'a}ndez-Lobato,
  Daniel, Bui, Thang, and Turner, Richard~E.
\newblock Black-box alpha divergence minimization.
\newblock In \emph{Proceedings of The 33rd International Conference on Machine
  Learning}, pp.\  1511--1520, 2016.

\bibitem[Hinton \& Van~Camp(1993)Hinton and Van~Camp]{hinton1993keeping}
Hinton, Geoffrey~E and Van~Camp, Drew.
\newblock Keeping the neural networks simple by minimizing the description
  length of the weights.
\newblock In \emph{COLT}, pp.\  5--13. ACM, 1993.

\bibitem[Hinton et~al.(2012)Hinton, Srivastava, Krizhevsky, Sutskever, and
  Salakhutdinov]{hinton2012improving}
Hinton, Geoffrey~E, Srivastava, Nitish, Krizhevsky, Alex, Sutskever, Ilya, and
  Salakhutdinov, Ruslan~R.
\newblock Improving neural networks by preventing co-adaptation of feature
  detectors.
\newblock \emph{arXiv preprint arXiv:1207.0580}, 2012.

\bibitem[Jordan et~al.(1999)Jordan, Ghahramani, Jaakkola, and
  Saul]{jordan1999introduction}
Jordan, Michael~I, Ghahramani, Zoubin, Jaakkola, Tommi~S, and Saul, Lawrence~K.
\newblock An introduction to variational methods for graphical models.
\newblock \emph{Machine learning}, 37\penalty0 (2):\penalty0 183--233, 1999.

\bibitem[Kalchbrenner \& Blunsom(2013)Kalchbrenner and
  Blunsom]{kalchbrenner2013recurrent}
Kalchbrenner, Nal and Blunsom, Phil.
\newblock Recurrent continuous translation models.
\newblock In \emph{EMNLP}, 2013.

\bibitem[Kendall \& Cipolla(2016)Kendall and Cipolla]{kendall2016modelling}
Kendall, Alex and Cipolla, Roberto.
\newblock Modelling uncertainty in deep learning for camera relocalization.
\newblock In \emph{2016 IEEE International Conference on Robotics and
  Automation (ICRA)}, pp.\  4762--4769. IEEE, 2016.

\bibitem[Kendall et~al.(2015)Kendall, Badrinarayanan, and
  Cipolla]{kendall2015bayesian}
Kendall, Alex, Badrinarayanan, Vijay, and Cipolla, Roberto.
\newblock Bayesian segnet: Model uncertainty in deep convolutional
  encoder-decoder architectures for scene understanding.
\newblock \emph{arXiv preprint arXiv:1511.02680}, 2015.

\bibitem[Krizhevsky et~al.(2012)Krizhevsky, Sutskever, and
  Hinton]{krizhevsky2012imagenet}
Krizhevsky, Alex, Sutskever, Ilya, and Hinton, Geoffrey~E.
\newblock Imagenet classification with deep convolutional neural networks.
\newblock In \emph{Advances in neural information processing systems}, pp.\
  1097--1105, 2012.

\bibitem[Kullback(1959)]{kullback1959information}
Kullback, Solomon.
\newblock \emph{Information theory and statistics}.
\newblock John Wiley \& Sons, 1959.

\bibitem[Kullback \& Leibler(1951)Kullback and
  Leibler]{kullback1951information}
Kullback, Solomon and Leibler, Richard~A.
\newblock On information and sufficiency.
\newblock \emph{The annals of mathematical statistics}, 22\penalty0
  (1):\penalty0 79--86, 1951.

\bibitem[LeCun \& Cortes(1998)LeCun and Cortes]{lecun1998mnist}
LeCun, Yann and Cortes, Corinna.
\newblock The mnist database of handwritten digits, 1998.

\bibitem[LeCun et~al.(1989)LeCun, Boser, Denker, Henderson, Howard, Hubbard,
  and Jackel]{lecun1989backpropagation}
LeCun, Yann, Boser, Bernhard, Denker, John~S, Henderson, Donnie, Howard,
  Richard~E, Hubbard, Wayne, and Jackel, Lawrence~D.
\newblock Backpropagation applied to handwritten zip code recognition.
\newblock \emph{Neural Computation}, 1\penalty0 (4):\penalty0 541--551, 1989.

\bibitem[LeCun et~al.(2006)LeCun, Chopra, Hadsell, Ranzato, and
  Huang]{lecun2006energy}
LeCun, Yann, Chopra, Sumit, Hadsell, Raia, Ranzato, M, and Huang, F.
\newblock A tutorial on energy-based learning.
\newblock \emph{Predicting structured data}, 1:\penalty0 0, 2006.

\bibitem[Li \& Turner(2016)Li and Turner]{li2016variational}
Li, Yingzhen and Turner, Richard~E.
\newblock R{\'e}nyi divergence variational inference.
\newblock In \emph{NIPS}, 2016.

\bibitem[Li et~al.(2015)Li, Hern{\'a}ndez-Lobato, and Turner]{li2015sep}
Li, Yingzhen, Hern{\'a}ndez-Lobato, Jos{\'e}~Miguel, and Turner, Richard~E.
\newblock Stochastic expectation propagation.
\newblock In \emph{Advances in Neural Information Processing Systems (NIPS)},
  2015.

\bibitem[MacKay(1992)]{mackay1992practical}
MacKay, David~JC.
\newblock A practical {B}ayesian framework for backpropagation networks.
\newblock \emph{Neural Computation}, 4\penalty0 (3):\penalty0 448--472, 1992.

\bibitem[Mikolov et~al.(2010)Mikolov, Karafi{\'a}t, Burget, {\v{C}}ernock{\`y},
  and Khudanpur]{mikolov2010recurrent}
Mikolov, Tom{\'a}{\v{s}}, Karafi{\'a}t, Martin, Burget, Luk{\'a}{\v{s}},
  {\v{C}}ernock{\`y}, Jan, and Khudanpur, Sanjeev.
\newblock Recurrent neural network based language model.
\newblock In \emph{Eleventh Annual Conference of the International Speech
  Communication Association}, 2010.

\bibitem[Minka(2005)]{minka2005divergence}
Minka, Tom.
\newblock Divergence measures and message passing.
\newblock Technical report, Microsoft Research, 2005.

\bibitem[Minka(2001)]{minka2001ep}
Minka, T.P.
\newblock {Expectation propagation for approximate Bayesian inference}.
\newblock In \emph{Conference on Uncertainty in Artificial Intelligence (UAI)},
  2001.

\bibitem[Minka(2004)]{minka2004powerep}
Minka, T.P.
\newblock {Power EP}.
\newblock Technical Report MSR-TR-2004-149, Microsoft Research, 2004.

\bibitem[Neal(1995)]{neal1995bayesian}
Neal, Radford~M.
\newblock \emph{Bayesian learning for neural networks}.
\newblock PhD thesis, University of Toronto, 1995.

\bibitem[Papernot et~al.(2016)Papernot, Goodfellow, Sheatsley, Feinman, and
  McDaniel]{papernot2016cleverhans}
Papernot, Nicolas, Goodfellow, Ian, Sheatsley, Ryan, Feinman, Reuben, and
  McDaniel, Patrick.
\newblock cleverhans v1.0.0: an adversarial machine learning library.
\newblock \emph{arXiv preprint arXiv:1610.00768}, 2016.

\bibitem[R{\'e}nyi(1961)]{renyi1961divergence}
R{\'e}nyi, Alfr{\'e}d.
\newblock On measures of entropy and information.
\newblock \emph{Fourth Berkeley symposium on mathematical statistics and
  probability}, 1, 1961.

\bibitem[Rumelhart et~al.(1985)Rumelhart, Hinton, and
  Williams]{rumelhart1985learning}
Rumelhart, David~E, Hinton, Geoffrey~E, and Williams, Ronald~J.
\newblock Learning internal representations by error propagation.
\newblock Technical report, DTIC Document, 1985.

\bibitem[Sennrich et~al.(2016)Sennrich, Haddow, and
  Birch]{sennrich2016Edinburgh}
Sennrich, Rico, Haddow, Barry, and Birch, Alexandra.
\newblock Edinburgh neural machine translation systems for wmt 16.
\newblock In \emph{Proceedings of the First Conference on Machine Translation},
  pp.\  371--376, Berlin, Germany, August 2016. Association for Computational
  Linguistics.

\bibitem[Srivastava et~al.(2014)Srivastava, Hinton, Krizhevsky, Sutskever, and
  Salakhutdinov]{srivastava2014dropout}
Srivastava, Nitish, Hinton, Geoffrey, Krizhevsky, Alex, Sutskever, Ilya, and
  Salakhutdinov, Ruslan.
\newblock Dropout: A simple way to prevent neural networks from overfitting.
\newblock \emph{The Journal of Machine Learning Research}, 15\penalty0
  (1):\penalty0 1929--1958, 2014.

\bibitem[Sundermeyer et~al.(2012)Sundermeyer, Schl{\"u}ter, and
  Ney]{sundermeyer2012lstm}
Sundermeyer, Martin, Schl{\"u}ter, Ralf, and Ney, Hermann.
\newblock {LSTM} neural networks for language modeling.
\newblock In \emph{INTERSPEECH}, 2012.

\bibitem[Sutskever et~al.(2014)Sutskever, Vinyals, and
  Le]{sutskever2014sequence}
Sutskever, Ilya, Vinyals, Oriol, and Le, Quoc~VV.
\newblock Sequence to sequence learning with neural networks.
\newblock In \emph{NIPS}, 2014.

\bibitem[Szegedy et~al.(2014)Szegedy, Liu, Jia, Sermanet, Reed, Anguelov,
  Erhan, Vanhoucke, and Rabinovich]{szegedy2014going}
Szegedy, Christian, Liu, Wei, Jia, Yangqing, Sermanet, Pierre, Reed, Scott,
  Anguelov, Dragomir, Erhan, Dumitru, Vanhoucke, Vincent, and Rabinovich,
  Andrew.
\newblock Going deeper with convolutions.
\newblock \emph{arXiv preprint arXiv:1409.4842}, 2014.

\bibitem[Turner \& Sahani(2011)Turner and Sahani]{turner2011two}
Turner, RE and Sahani, M.
\newblock Two problems with variational expectation maximisation for
  time-series models.
\newblock \emph{Inference and Estimation in Probabilistic Time-Series Models},
  2011.

\bibitem[Van~Erven \& Harremo{\"e}s(2014)Van~Erven and
  Harremo{\"e}s]{van_erven2014renyi}
Van~Erven, Tim and Harremo{\"e}s, Peter.
\newblock R{\'e}nyi divergence and {Kullback-Leibler} divergence.
\newblock \emph{Information Theory, IEEE Transactions on}, 60\penalty0
  (7):\penalty0 3797--3820, 2014.

\bibitem[Wan et~al.(2013)Wan, Zeiler, Zhang, LeCun, and
  Fergus]{wan2013regularization}
Wan, L, Zeiler, M, Zhang, S, LeCun, Y, and Fergus, R.
\newblock Regularization of neural networks using dropconnect.
\newblock In \emph{ICML-13}, 2013.

\bibitem[Yang et~al.(2016)Yang, Kwitt, and Niethammer]{yang2016fast}
Yang, Xiao, Kwitt, Roland, and Niethammer, Marc.
\newblock Fast predictive image registration.
\newblock \emph{arXiv preprint arXiv:1607.02504}, 2016.

\end{thebibliography}
}
\bibliographystyle{icml2017}

\appendix
\onecolumn

\section{Code Example}



The following is a code snippet showing how our inference can be implemented with a few lines of Keras code \citep{keras2015}. We define a new loss function \texttt{bbalpha\_softmax\_cross\_entropy\_with\_mc\_logits}, that takes MC sampled logits as an input. This is demonstrated for the case of classification. Regression can be implemented in a similar way.

{\small
\begin{minted}{python}
def bbalpha_softmax_cross_entropy_with_mc_logits(alpha):
  def loss(y_true, mc_logits):
    # mc_logits: output of GenerateMCSamples, of shape M x K x D
    mc_log_softmax = mc_logits - K.max(mc_logits, axis=2, keepdims=True)
    mc_log_softmax = mc_log_softmax - logsumexp(mc_log_softmax, 2)
    mc_ll = K.sum(y_true * mc_log_softmax, -1)  # M x K
    return - 1. / alpha * (logsumexp(alpha * mc_ll, 1) + K.log(1.0 / K_mc))
  return loss
\end{minted}
}


MC samples for this loss can be generated using \texttt{GenerateMCSamples}, with \texttt{layers} being a list of Keras initialised layers:

{\small
\begin{minted}{python}
def GenerateMCSamples(inp, layers, K_mc=20):
  output_list = []
  for _ in xrange(K_mc):
    output_list += [apply_layers(inp, layers)]
  def pack_out(output_list):
    output = K.pack(output_list) # K_mc x nb_batch x nb_classes
    return K.permute_dimensions(output, (1, 0, 2)) # nb_batch x K_mc x nb_classes
  def pack_shape(s):
    s = s[0]
    return (s[0], K_mc, s[1])
  out = Lambda(pack_out, output_shape=pack_shape)(output_list)
  return out
\end{minted}
}

The above two functions rely on the following auxiliary functions:
{\small
\begin{minted}{python}
def logsumexp(x, axis=None):
  x_max = K.max(x, axis=axis, keepdims=True)
  return K.log(K.sum(K.exp(x - x_max), axis=axis, keepdims=True)) + x_max
  
def apply_layers(inp, layers):
  output = inp
  for layer in layers:
      output = layer(output)
  return output
\end{minted}
}

\section{Alpha-divergence minimisation}
\label{sec:divergence_def}
There are various available definitions of $\alpha$-divergences, and in this work we mainly used two of them: Amari's definition \cite{amari1985book} adapted to EP context \cite{minka2005divergence}, and R{\'e}nyi divergence \cite{renyi1961divergence} which is more used in information theory research.
\begin{itemize}
	\item Amari's $\alpha$-divergence \cite{amari1985book}:
	$$\mathrm{D}_{\alpha}[p||q] = \frac{1}{\alpha (1 - \alpha) } \left( 1 - \int p(\bo)^{\alpha} q(\bo)^{1 - \alpha} d\bo \right).$$
	\item R{\'e}nyi's $\alpha$-divergence \cite{renyi1961divergence}:
	$$\mathrm{R}_{\alpha}[p||q] = \frac{1}{\alpha - 1} \log \int p(\bo)^{\alpha} q(\bo)^{1 - \alpha} d\bo.$$
\end{itemize}
These two divergence can be converted to each other, e.g. $\mathrm{D}_{\alpha}[p||q] = \frac{1}{\alpha (1 - \alpha)} \left( 1 - \exp \left[(\alpha - 1) \mathrm{R}_{\alpha}[p||q] \right] \right)$.
In power EP \cite{minka2004powerep}, this $\alpha$-divergence is minimised using projection-based updates.  When the approximate posterior $q$ has an exponential family form, minimising $\mathrm{D}_{\alpha}[p||q]$ requires moment matching to the ``tilted distribution'' $\tilde{p}_{\alpha}(\bo) \propto p(\bo)^{\alpha} q(\bo)^{1-\alpha}$. This projection update might be intractable for non-exponential family $q$ distributions, and instead BB-$\alpha$ deploys a gradient-based update to search a local minimum. We will present the original derivation of the BB-$\alpha$ energy below and discuss how it relates to power EP.

\section{Original Derivation of BB-$\alpha$ Energy}
Here we include the original formulation of the BB-$\alpha$ energy for completeness. Consider approximating a distribution of the following form
$$p(\bo) = \frac{1}{Z} p_0(\bo) \prod_{n}^{N} f_n(\bo),$$
in which the prior distribution $p_0(\bo)$ has an exponential family form
$p_0(\bo) \propto \exp \left[ \lambda_0^{T}\phi(\bo) \right]$. Here $\lambda_0$ is called natural parameter or canonical parameter of the exponential family distribution, and $\phi(\bo)$ is the sufficient statistic. As the factors $f_n$ might not be conjugate to the prior, the exact posterior no longer belongs to the same exponential family as the prior, and hence need approximations. EP construct such approximation by first approximating each complicated factor $f_n$ with a simpler one $\tilde{f}_n(\bo) \propto \exp \left[ \lambda_n^{T}\phi(\bo) \right]$, then constructing the approximate distribution as $$q(\bo) = \frac{1}{Z(\lambda_q)} \exp \left[ \left( \sum_{n=0}^{N} \lambda_n \right)^{T}\phi(\bo) \right],$$ 
with $\lambda_q = \lambda_0 + \sum_{n=1}^N \lambda_n$ and $Z(\lambda_q)$ the normalising constant/partition function. These \emph{local} parameters are updated using the following procedure (for $\alpha \neq 0$):
\begin{itemize}
\item[1] compute cavity distribution $q^{\setminus n}(\bo) \propto q(\bo) / f_n(\bo)$, equivalently. $\lambda^{\setminus n} \leftarrow \lambda_q -  \lambda_n$;
\item[2] compute the tilted distribution by inserting the likelihood term $\tilde{p}_n(\bo) \propto q^{\setminus n}(\bo) f_n(\bo)$;
\item[3] compute a projection update: $\lambda_q \leftarrow \argmin_{\lambda} \mathrm{D}_{\alpha}[\tilde{p}_n||q_{\lambda}]$ with $q_{\lambda}$ an exponential family with natural parameter $\lambda$;
\item[4] recover the site approximation by $\lambda_n \leftarrow \lambda_q - \lambda^{\setminus n}$ and form the final update $\lambda_q \leftarrow \sum_n \lambda_n + \lambda_0$.
\end{itemize}

When converged, the solutions of $\lambda_n$ return a fixed point of the so called \emph{power EP energy}:
\begin{equation}
\mathcal{L}_{\text{PEP}}(\lambda_0,\{ \lambda_n \}) =  \log Z(\lambda_0) + (\frac{N}{\alpha}-1) \log Z(\lambda_q) 
  - \frac{1}{\alpha} \sum_{n=1}^{N} \log \int f_n(\bo)^{\alpha} \exp\left[ (\lambda_q - \alpha \lambda_n)^T \phi(\bo) \right] d\bo.
\label{eq:energy}
\end{equation}
But more importantly, before convergence all these local parameters $\lambda_n$ are maintained in memory. This indicates that power EP does not scale with big data: consider Gaussian approximations which has $\mathcal{O}(d^2)$ parameters with $d$ the dimensionality of $\bo$. Then the space complexity of power EP is $\mathcal{O}(Nd^2)$, which is clearly prohibitive for big models like neural networks that are typically applied to large datasets.
BB-$\alpha$ provides a simple solution of this memory overhead by sharing the local parameters, i.e. defining $\lambda_n = \lambda$ for all $n = 1, ..., N$. Furthermore, under the mild condition that the exponential family is regular, there exist a one-to-one mapping between $\lambda_q$ and $\lambda$ (given a fixed $\lambda_0$). Hence we arrive at a ``global'' optimisation problem in the sense that only one parameter $\lambda_q$ is optimised, where the objective function is the BB-$\alpha$ energy
\begin{equation}
\mathcal{L}_{\alpha}(\lambda_0, \lambda_q) =  \log Z(\lambda_0) - \log Z(\lambda_q) 
  - \frac{1}{\alpha} \sum_{n=1}^{N} \log \mathbb{E}_{q} \left[ \left( \frac{f_n(\bo)}{\exp \left[ \lambda^T \phi(\bo) \right] } \right)^{\alpha} \right].
\label{eq:energy}
\end{equation}
One could verify that this is equivalent to the BB-$\alpha$ energy function presented in the main text by considering exponential family $q$ distributions.

Although empirical evaluations have demonstrated the superior performance of BB-$\alpha$, the original formulation is difficult to interpret for practitioners. First the local alpha-divergence minimisation interpretation is inherited from power EP, and the intuition of power EP itself might already pose challenges for practitioners. Second, the derivation of BB-$\alpha$ from power EP is ad hoc and lacks theoretical justification. It has been shown that power EP energy can be viewed as the dual objective to a continuous version of Bethe free-energy, in which $\lambda_n$ represents the Lagrange multiplier of the constraints in the primal problem. Hence tying the Lagrange multipliers would effectively changes the primal problem, thus losing a number of nice guarantees. Nevertheless this approximation has been shown to work well in real-world settings, which motivated our work to extend BB-$\alpha$ to dropout approximation.

\section{Full Regression Results}
\begin{table}[ht]
\centering
\caption{Regression experiment: Average negative test log likelihood/nats}
\label{tab:reg_results_ll}
\begin{tabular}{l@{\ica}r@{\ica}r@{\ica}r@{$\pm$}l@{\ica}r@{$\pm$}l@{\ica}r@{$\pm$}l@{\ica}r@{$\pm$}l@{\ica}r@{$\pm$}l@{\ica}r@{$\pm$}l@{\ica}}
\hline
\bf{Dataset}&{N}&{D}&\multicolumn{2}{c}{\bf{$\alpha=0.0$}}&\multicolumn{2}{c}{\bf{$\alpha=0.5$}}&\multicolumn{2}{c}{\bf{$\alpha=1.0$}}&\multicolumn{2}{c}{\bf{HMC}}&\multicolumn{2}{c}{\bf{GP}}&\multicolumn{2}{c}{\bf{VI-G}}\\
\hline
boston&506&13&2.42&0.05&2.38&0.06&2.50&0.10&2.27&0.03&2.22&0.07&2.52&0.03\\
concrete&1030&8&2.98&0.02&2.88&0.02&2.96&0.03&2.72&0.02&2.85&0.02&3.11&0.02\\
energy&768&8&1.75&0.01&0.74&0.02&0.81&0.02&0.93&0.01&1.29&0.01&0.77&0.02\\
kin8nm&8192&8&-0.83&0.00&-1.03&0.00&-1.10&0.00&-1.35&0.00&-1.31&0.01&-1.12&0.01\\
power&9568&4&2.79&0.01&2.78&0.01&2.76&0.00&2.70&0.00&2.66&0.01&2.82&0.01\\
protein&45730&9&2.87&0.00&2.87&0.00&2.86&0.00&2.77&0.00&2.95&0.05&2.91&0.00\\
red wine&1588&11&0.92&0.01&0.92&0.01&0.95&0.02&0.91&0.02&0.67&0.01&0.96&0.01\\
yacht&308&6&1.38&0.01&1.08&0.04&1.15&0.06&1.62&0.01&1.15&0.03&1.77&0.01\\
naval&11934&16&-2.80&0.00&-2.80&0.00&-2.80&0.00&-7.31&0.00&-4.86&0.04&-6.49&0.29\\
year&515345&90&3.59&NA&3.54&NA&-3.59&NA&NA&NA&0.65&NA&3.60&NA\\
\hline
\end{tabular}
\end{table}

\begin{table}[ht]
\centering
\caption{Regression experiment: Average test RMSE}
\label{tab:reg_results_ll}
\begin{tabular}{l@{\ica}r@{\ica}r@{\ica}r@{$\pm$}l@{\ica}r@{$\pm$}l@{\ica}r@{$\pm$}l@{\ica}r@{$\pm$}l@{\ica}r@{$\pm$}l@{\ica}r@{$\pm$}l@{\ica}}
\hline
\bf{Dataset}&{N}&{D}&\multicolumn{2}{c}{\bf{$\alpha=0.0$}}&\multicolumn{2}{c}{\bf{$\alpha=0.5$}}&\multicolumn{2}{c}{\bf{$\alpha=1.0$}}&\multicolumn{2}{c}{\bf{HMC}}&\multicolumn{2}{c}{\bf{GP}}&\multicolumn{2}{c}{\bf{VI-G}}\\
\hline
boston&506&13&2.85&0.19&2.97&0.19&3.04&0.17&2.76&0.20&2.43&0.07&2.89&0.17\\
concrete&1030&8&4.92&0.13&4.62&0.12&4.76&0.15&4.12&0.14&5.55&0.02&5.42&0.11\\
energy&768&8&1.02&0.03&1.11&0.02&1.10&0.02&0.48&0.01&1.02&0.02&0.51&0.01\\
kin8nm&8192&8&0.09&0.00&0.09&0.00&0.08&0.00&0.06&0.00&0.07&0.00&0.08&0.00\\
power&9568&4&4.04&0.04&4.01&0.04&3.98&0.04&3.73&0.04&3.75&0.03&4.07&0.04\\
protein&45730&9&4.28&0.02&4.28&0.04&4.23&0.01&3.91&0.02&4.83&0.21&4.45&0.02\\
red wine&1588&11&0.61&0.01&0.62&0.01&0.63&0.01&0.63&0.01&0.57&0.01&0.63&0.01\\
yacht&308&6&0.76&0.05&0.85&0.06&0.88&0.06&0.56&0.05&1.15&0.09&0.81&0.05\\
naval&11934&16&0.01&0.00&0.01&0.00&0.01&0.00&0.00&0.00&0.00&0.00&0.00&0.00\\
year&515345&90&8.66&NA&8.80&NA&8.97&NA&NA&NA&0.79&NA&8.88&NA\\
\hline
\end{tabular}
\end{table}

\end{document}